%% file: main.tex
\documentclass[10pt,twocolumn,letterpaper]{article}

 \usepackage{cvpr}              %

\input{preamble}

\definecolor{cvprblue}{rgb}{0.21,0.49,0.74}
\usepackage[pagebackref,breaklinks,colorlinks,allcolors=cvprblue]{hyperref}

\def\paperID{12463} %
\def\confName{CVPR}
\def\confYear{2025}

\usepackage[toc,page,header]{appendix}
\usepackage{minitoc}

\title{Composing Parts for Expressive Object Generation} %

\author{%
     Harsh Rangwani\textsuperscript{1} \quad Aishwarya Agarwal\textsuperscript{1} \quad Kuldeep Kulkarni\textsuperscript{1} \\ \quad R. Venkatesh Babu\textsuperscript{2}  \quad 
     Srikrishna Karanam\textsuperscript{1} \\ \\
     \textsuperscript{1}Adobe Research \quad
      \textsuperscript{2} Indian Institute of Science
 }

\begin{document}
\doparttoc %
\faketableofcontents %

\twocolumn[{%
\renewcommand\twocolumn[1][]{#1}%
\maketitle

\begin{center}
 \centering
 \captionsetup{type=figure}
 \vspace{-5pt}
 \includegraphics[width=\textwidth]{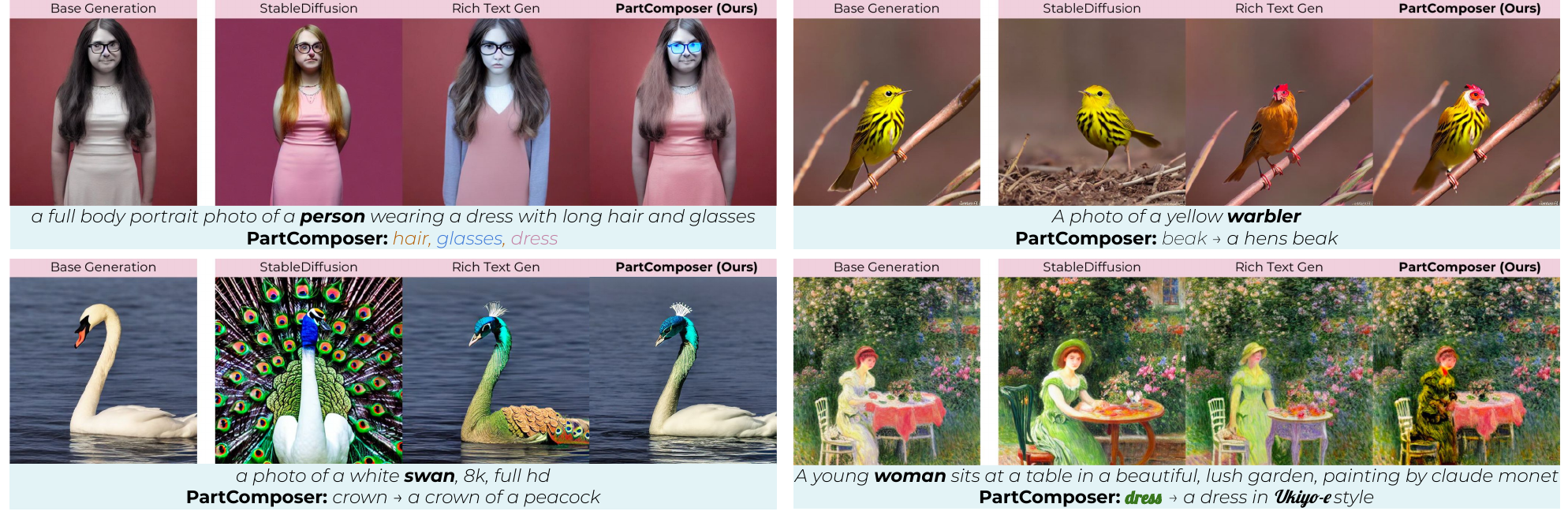}
 \vspace{-5mm}
 \caption{\textbf{Base Generation (left) Comparison with Methods for Generations with Parts Details (right).}  PartComposer allows the generation of object images with specified attributes (color, style \etc) of parts for the \textbf{chosen} $\textbf{object}$ in the base text prompt. StableDiffusion and Rich-Text~\cite{ge2023expressive} methods with part details either ignore the part instructions or generate inconsistent ojects.  }

 \label{fig:teaser_qual}
\end{center}%
}]
\setcounter{footnote}{0} 
\begin{abstract}
 Image composition and generation are processes where the artists need control over various parts of the generated images. However, the current state-of-the-art generation models, like Stable Diffusion, cannot handle fine-grained part-level attributes in the text prompts. Specifically, when additional attribute details are added to the base text prompt, these text-to-image models either generate an image vastly different from the image generated from the base prompt or ignore the attribute details. To mitigate these issues, we introduce PartComposer, a training-free method that enables image generation based on fine-grained part-level attributes specified for objects in the base text prompt. This allows more control for artists and enables novel object compositions by combining distinctive object parts. PartComposer first localizes object parts by denoising the object region from a specific diffusion process. This enables each part token to be localized to the right region. After obtaining part masks, we run a localized diffusion process in each part region based on fine-grained part attributes and combine them to produce the final image. All stages of PartComposer are based on repurposing a pre-trained diffusion model, which enables it to generalize across domains. We demonstrate the effectiveness of part-level control provided by PartComposer through qualitative visual examples and quantitative comparisons with contemporary baselines. 
\end{abstract}

\section{Introduction}
\input{paper-sections/intro}

\section{Related Works}
\input{paper-sections/related-works}

\section{Method}

\input{paper-sections/method}
\section{Experimental Analysis}
\input{paper-sections/experimental-analysis}

\section{Conclusion}
\input{paper-sections/conclusion}

\input{paper-sections/supp}

\newpage
{
    \small
    \bibliographystyle{ieeenat_fullname}
    \bibliography{main}
}

\end{document}

%% file: preamble.tex
\usepackage[dvipsnames]{xcolor}

\usepackage{adjustbox} %
\usepackage{url}            %
\usepackage{booktabs}       %
\usepackage{amsfonts}       %
\usepackage{nicefrac}       %
\usepackage{microtype}      %
\usepackage{multirow}
\usepackage{multicol}
\usepackage{dsfont}
\usepackage{wrapfig,lipsum}
\usepackage{breakurl}
\usepackage{adjustbox}
\usepackage{amsmath}

\newcommand{\indc}{\mathds{1}}

\usepackage{soul}
\definecolor{Gray}{gray}{0.95}
\definecolor{gray_1}{HTML}{e7edee}
\definecolor{gray_2}{HTML}{cfd4d5}
\definecolor{gray_3}{HTML}{b7bbbc}
\definecolor{gray_4}{HTML}{a2a5a6}
\definecolor{b_1}{HTML}{d2e2df}
\definecolor{b_2}{HTML}{e1ebe9}
\definecolor{b_3}{HTML}{e9f2f1}
\definecolor{b_4}{HTML}{f3faf9}
\newcommand{\bp}{\mathbf{p}}
\newcommand{\ba}{\mathbf{a}}

\newcommand{\bm}{\mathbf{m}}
\newcommand{\bM}{\mathbf{M}}

\usepackage{dsfont}

\DeclareMathOperator{\argmax}{argmax}

%% file: paper-sections/intro.tex
\label{sec:intro}
Image generation with large generative diffusion models like StableDiffusion~\cite{rombach2022high}, DALLE~\cite{ramesh2021zero}, etc., has become prevalent due to their superior quality and extensive world knowledge. These models are trained on large image-caption datasets and are trained to generate images based on a given text prompt (description). As image composition and creation are creative processes where the artists need control over various parts of the image being generated. However, adding additional details for controlling part appearance in the text prompt either changes the generated image entirely or ignores the part instructions~\cite{chefer2023attend} (Fig.~\ref{fig:teaser_qual}). 

Various works aim to provide improved spatial controls to image generation, as they allow image generation conditioned on segmentation masks ~\cite{bar2023multidiffusion}, edge maps~\cite{zhang2023adding}, bounding boxes~\cite{chen2024training, li2023gligen}  etc. Popular methods such as ControlNet~\cite{zhang2023adding}, GLIGEN~\cite{li2023gligen}, etc. require specifying training of these conditional modules, which allows for the controllability of these large generative models. Further, there has been some development of training-free approaches~\cite{chefer2023attend, chen2024training} which enable controlled generation of objects by modulating the internal cross-attention activations in the diffusion process. This demonstrates that these pre-trained generative models contain information about the spatial parts of the image, and modulating them effectively can lead to image compositions. 

However, despite several attempts ~\cite{bhat2023loosecontrol, zhang2023adding}, the semantic controllability of the image generation is still restricted to specifying details at the object level. The object level details can be restrictive, as often creative designers synthesize object parts (\eg shirt, trouser etc.) and then compose them~\cite{ko2023large}. Further, variations in semantic parts are often very distinctive and are uniquely used to identify objects~\cite{ma2023directed, choudhury2021unsupervised}. For example, we usually identify species of birds by their unique \emph{beak}. Due to this, semantic understanding and recognition of parts have been widely studied as a topic in computer vision~\cite{choudhury2021unsupervised, maji2013fine, hung2019scops, collins2018deep}. Hence, providing controllability for image synthesis at the object semantic part level can enable a large variety of image compositions~\cite{ko2023large}. Towards this goal, we introduce \textbf{PartComposer}, which provides users with an interface through which they can select the object in the scene and provide a semantic part-level description with fine-grained details for the object generation.

In \textbf{PartComposer}, we develop a scheme to extract part-level localization masks from the Diffusion model. We introduce a parallel part diffusion process that generates masks for the object parts. The core idea of the approach is \emph{that by forcing the part diffusion model to specifically denoise only the object region in the image, it is possible to understand the locations of various parts in the object}. After denoising just the object through the part model, the information inside the part diffusion model present in attention maps can be used to generate the masks for various object parts. 
In the following Part Generation stage, PartComposer utilizes the part masks and attributes of each part the user provides. We enable users to provide a highly expressive specification of parts by using a Rich-Text~\cite{ge2023expressive} interface, which allows the specifying attributes, such as style, color, etc., for each part. For the final image generation, taking inspiration from recent studies like Rich-Text Generation~\cite{ge2023expressive}, Multi-Diffusion~\cite{bar2023multidiffusion} etc., we compose the various object parts, by running parallel masked diffusion process for each part while combining them periodically into the image. This combination enables harmonious composition of parts, and also masking ensures only local modifications to the regions corresponding to each part of the object specified. In the PartComposer method, we only use the pre-trained StableDiffusion model, making it a generalizable and training-free approach.

We extensively test the proposed PartComposer approach for the zero-shot object part segmentation, which is a challenging setup in computer vision. We evaluate the part segmentation approach on DeepFashion~\cite{liuLQWTcvpr16DeepFashion} and CUB200~\cite{WahCUB_200_2011} datasets, where our method significantly outperforms the baseline StableDiffusion (SD). Further, to evaluate the PartComposer image composition abilities, we also provide quantitative results along with a user preference study, where our method significantly outperforms the baselines in generating images consistent with the described parts ({Fig.~\ref{fig:teaser_qual}}).
We summarize the core contributions of our paper below:  
\begin{enumerate}[topsep=0pt,itemsep=-1ex,partopsep=1ex,parsep=1ex]

    \item We introduce PartComposer, a training-free method that enables the generation of object images by using provided fine-grained details for the parts of the object. For example, while generating a bird image we can specify a detailed description of its beak.
    \item In PartComposer, we introduce a novel Part-Diffusion process, which localizes and provides masks for parts of a base object generated by the Diffusion model (Fig.~\ref{fig:localization}). To localize object parts, we introduce a novel segmentation scheme that uses the attention maps of the base diffusion and part diffusion process to obtain accurate masks for localized parts.
    \item In PartComposer, after localization, we enable the generation of parts from the pre-trained diffusion model based on the Rich-Text description for the parts provided by the user. PartComposer then composes the image to harmoniously blend all object parts with the background by combining localized diffusion paths (Fig.~\ref{fig:overview}). 
\end{enumerate}

%% file: paper-sections/related-works.tex
 \noindent \textbf{Text-to-Image Generative Models.}
 \begin{figure*}[!t]
    \centering

    \includegraphics[width=0.9\textwidth,height=6.5cm]{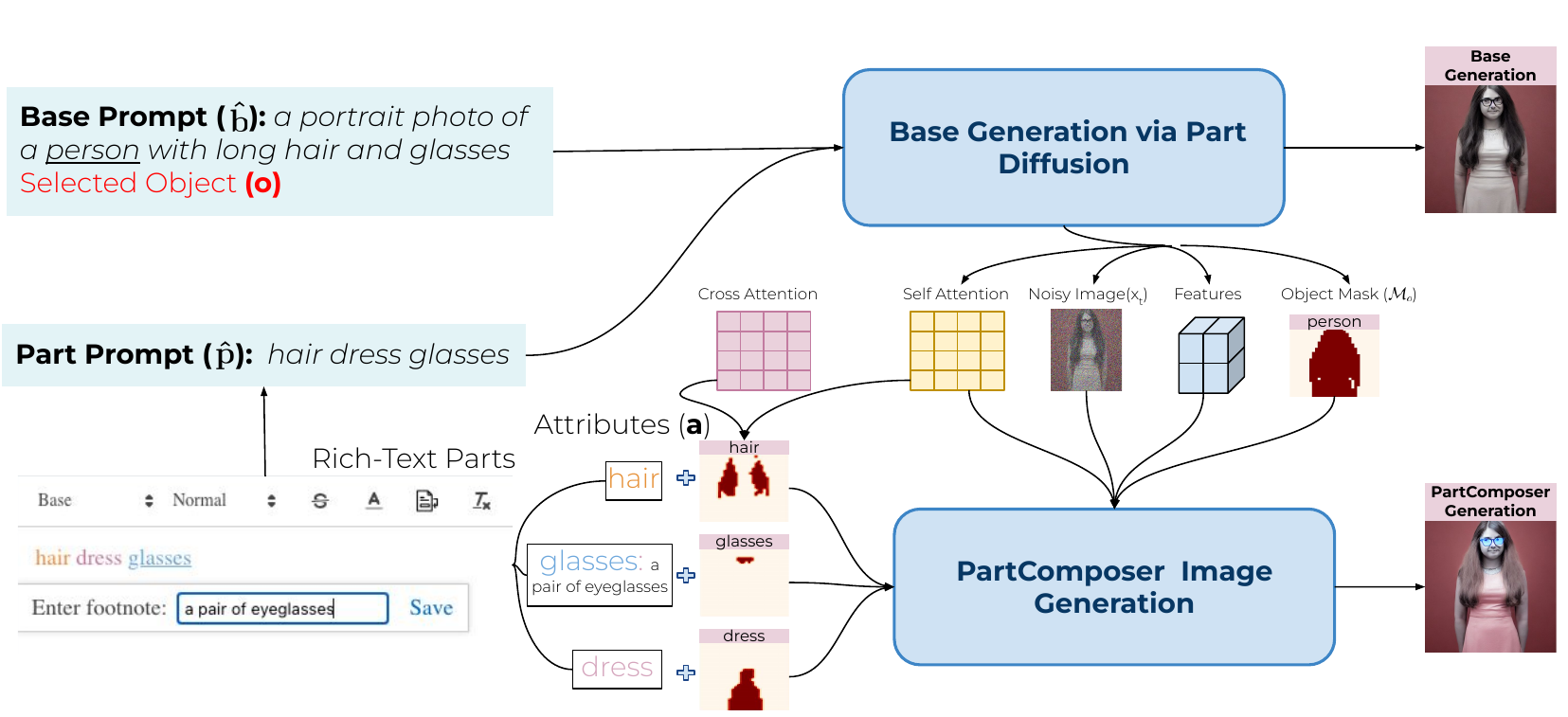}
        \vspace{-3.5mm}

    \caption{\textbf{PartComposer} takes the input of a base prompt, the selected object $\mathbf{o}$, and a Rich-Text description (\,  i.e., footnote, colors, \, etc.) of the parts. The Part Diffusion process generates the object masks for the specified parts. Then, PartComposer runs a parallel Region Diffusion process to generate attributes of the specified parts, guided by base generation's intermediate outputs. }
    \vspace{-5mm}
    \label{fig:overview}
    
\end{figure*}
The text-to-image models synthesize images by following a textual description provided as a prompt. These models have recently become mainstream due to their superior image generation quality and significant knowledge base. This is an outcome of the availability of large-scale image caption datasets~\cite{schuhmann2022laion} and highly parallelized GPU clusters. Almost all kinds of generative models, such as GANs~\cite{kang2023scaling}, Autoregressive~\cite{yu2022scaling, saharia2022photorealistic}, and Diffusion models~\cite{ramesh2021zero, rombach2022high}, have shown significant improvements in quality with training on these large image caption datasets. Among these models, the StableDiffusion (SD)~\cite{rombach2022high} models, based on denoising diffusion in latent space are popular due to their open-source nature, which we also utilize for experiments in our work. 

 \noindent \textbf{Text-to-Image Models for Downstream Tasks.} Generative models, in general, have been useful for various downstream tasks, particularly ones based on per-pixel prediction like Segmentation~\cite{abdal2021labels4free, choi2019self}, Depth Estimation~\cite{bhattad2024stylegan}, etc. As layers near the image generation output have features that capture the pixel-pixel relation~\cite{bhattad2024stylegan}. With the large-scale text-to-image generative models, these models often perform very competitively~\cite{xu2023odise, tian2023diffuse} to discriminative methods, on tasks like segmentation. However, one commonality among most of these segmentation methods is that they operate at the granularity of object or instance level. Text-to-image models here are at an advantage as the usual image captions describe the scene at an object level. In this work, we take a step further in exploring the object part-level knowledge of these text-to-image diffusion models.

 \noindent \textbf{Part Discovery and Segmentation.}
Part Discovery and Attribution were an integral part of computer vision pipelines classically, as these approaches were robust to viewpoint variations~\cite{maji2013fine, Vedaldi_2014_CVPR, DBLP:conf/bmvc/LazebnikSP04}. In deep learning, the unsupervised (self-supervised) approaches for part discovery like SCOPS~\cite{hung2019scops} and Unsup-Parts~\cite{choudhury2021unsupervised} became popular as they generalize to object parts across categories. In this work, we go one step ahead and operate in a zero-shot unsupervised part segmentation setting, generating part masks from T2I models.

 \noindent \textbf{Controllable Image Generation.} Masks, bounding boxes, edge maps, depth maps, etc., have been explored to control the generations of text-to-image diffusion models~\cite{li2023gligen, zhang2023adding, bhat2023loosecontrol, dahary2024yourself, parihar2024precisecontrol,parihar2024text2place} in addition to text. Further, various other approaches~\cite{alaluf2023cross, tewel2024training, voynov2023p+, agarwal2023astar, chefer2023attend, agarwal2024training, wang2023tokencompose, parihar2024balancing} achieve control of image semantics through modulating diffusion models. Despite this, the text-to-image generation control at the object part level is under-explored; a recent work~\cite{ng2024partcraft} tries to do it in a controlled supervised setup using part-masks. Contrary to that, in our work PartComposer we explore a generalized training-free zero-shot setting.

%% file: paper-sections/method.tex
\begin{figure*}[t]
    \centering
    \includegraphics[width=\textwidth]{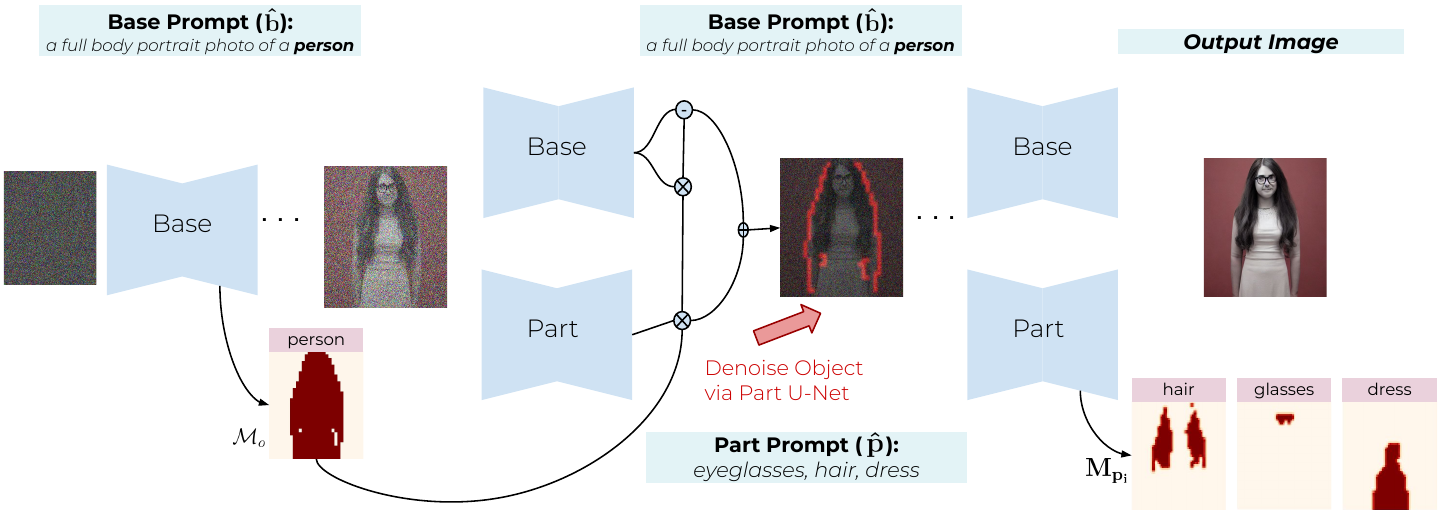}
    \vspace{-3mm}
    \caption{\textbf{Localization.} We obtain the mask ($\mathcal{M}_{o}$) for object in middle of diffusion process. We then denoise object in masked region, using parts $\bp_i$ conditioned U-Net. Due to part denoising, the attention maps of $\bp_i$ can now localize parts as masks $\mathbf{M}_{\bp_i}$.}
    \vspace{-5mm}
    \label{fig:localization}
\end{figure*}

In this section, we introduce PartComposer, our method to synthesize objects based on the description of parts of the objects. In PartComposer, we ask the user to specify a base prompt and the token for the object for which it wants to synthesize parts. Then, we provide the user a \textit{rich-text}~\cite{ge2023expressive} editor (Fig.~\ref{fig:overview}) to specify the parts and their description. PartComposer involves two diffusion steps, \textbf{a) Part Localization:} In the early diffusion stage, we get a mask for the object we want to divide into parts. Then, we perform denoising in later stage from a U-Net condition on parts to fill the masked region of the object, during which it learns to denoise different object parts. Due to this, the attention maps for various parts highlight the correct part region, which we use to extract the part mask. The infilling process is the major contribution of the PartComposer method (Fig,~\ref{fig:localization}).  
\textbf{b) Part Generation: } For generating parts, we combine region-specific diffusion processes for various parts by iteratively merging them inspired by MuliDiffusion~\cite{bar2023multidiffusion}, Rich-Text Generation~\cite{ge2023expressive} etc. However, till now, most of these works have combined the diffusion process for generating objects; for the first time, we have demonstrated its effectiveness in generating object parts. We provide an overview of the PartComposer pipeline in Fig.~\ref{fig:overview}.

\subsection{Problem Setup}
We consider each span of tokens $\mathbf{p}_i$ as indicative of describing one part of the object, with attributes $\ba_i$ describing its overall appearance. Our design choices are based on the rich-text generation~\cite{ge2023expressive}, and we allow the user to specify the following type of attributes for part generation (Fig.~\ref{fig:overview}): 

 \noindent \textbf{Part Description (\ie footnote).} It is an important attribute that specifies the part of an object; for example, for the part ($\bp_i$) `crown' of a bird, we can specify the attribute ($\ba_i^{p}$ = `a crown of a peacock'). This helps specify novel part descriptions, which can lead to artistic novel compositions. 

 \noindent \textbf{Font Color.} It helps us specify the exact RGB values for the color attribute $\ba_i^{c}$ we want for the object part $\bp_i$. The exact value of RGB allows fine-grained control over the color of the desired part, whereby just specifying specific colors like `brick red' leads to ignorance by Stable Diffusion~\cite{ge2023expressive}.
    
\noindent \textbf{Font Style.} The font style is indicative of the specific artistic style $\ba_{i}^{s}$ like `of Claude Monet' and `of Van Gogh' when synthesizing images of paintings. This instructs the model to generate a part following a specific style for the given prompt and blend it with other parts of the object (Sec.~\ref{sec:analysis}).

\noindent \textbf{Font Size.} The font size controls the relative size of each part in a generation~\cite{hertz2022prompt}. We use the $\ba^{w}_{i}$ to denote size.

\subsection{Part Localization}
\label{subsec:part_loc}

For part localization, we first run the base stable diffusion model for the given text prompt and extract the token map $\mathcal{M}_{o}$ for the object $\mathbf{o}$ specified by the user, by following the technique of clustering the self-attention maps~\cite{patashnik2023localizing, ge2023expressive}. We obtain the mask using this clustering after denoising between the initial step ($T$) until a threshold time step $T_{th}$. After obtaining the binary object mask $\mathcal{M}_{o}$ using attention masks~\cite{patashnik2023localizing, ge2023expressive}, we run two parallel diffusion processes where one contains input from the base prompt and the other contains input from the part prompts $\hat{\bp} = [\bp_1, \bp_2, ..., \bp_n]$, a token $\bp_i$ for each of the parts specified in the part text prompt. We denote the $\hat{\mathbf{b}}$ to denote the base text prompt and  $\hat{\bp}$ to denote the part prompt. For the time $t \leq T_{th}$ (by default, we use $T_{th} \approx T/2$), we denoise the U-Net in the object region by denoising it with both the combination of the Part Prompt output and Base Prompt output. Due to this, the Part-Based U-Net gets the information regarding various parts in the object (Fig.~\ref{fig:localization}). We now mathematically define the output noise $\epsilon_{t}$ for the diffusion process with Part-Diffusion below:

\begin{equation}
    \epsilon_{t} = \alpha \mathcal{M}_{o} \odot D(x_t, \hat{\bp}, t) + (1 - \alpha \mathcal{M}_{o}) \odot D(x_t, \hat{\mathbf{b}}, t)
    \label{eq:alpha-mask}
\end{equation}

here, the $\alpha$ is the hyper-parameter controlling the strength of the part prompt diffusion output, and $D$ is the output of the pre-trained U-Net of a text-to-image diffusion model. The above denoising process is followed for t steps until the last step to produce a base image corresponding to the given prompt $\hat{b}$. Keeping a high $T_{th}$ and low $\alpha$, makes minimal changes in output as if the denoising diffusion was done with original prompt $\hat{b}$ (See Suppl. Sec.). With this part denoising of object, we obtain attention masks, from which we extract the localization information of part $\bp_{t}$. We now describe the process of obtaining the part masks.

\noindent \textbf{Token Maps for Parts.} We first take the part tokens $\hat{\bp}$, which are a concatenation of the part names (\ie `beak crown wings'), which may not make a meaningful text prompt. Hence, we initialize text embeddings for all these tokens $\bp_i$ by passing the meaningful text prompt having the following template: `` A photo of $\bp_i$ of a $\mathbf{o}$ '' where $\bp_{i}$ is the object part name and $\mathbf{o}$ is the object name. This serves two purposes: first, it makes the text embedding meaningful, and second, it introduces some invariance from the order of part specification in part prompt. These embeddings are then passed as text embeddings to the Part U-Net for denoising. After running the denoising process, we aggregate the self-attention maps across multi-heads and time steps (from 32 $\times$ 32 resolution) for both the base and part U-Net diffusion branches, taking inspiration from works that demonstrate that attention can localize objects~\cite{chefer2023attend, ge2023expressive, tang2022daam, ma2023directed}. We then perform spectral clustering on these attention maps to form 
 k segmentation maps $\hat{\bM}$ (32 $\times$ 32), based on pixel-pixel similarity. To attribute these K-segments to the part specified by the user, we aggregate the cross attention of Part U-Net diffusion process. For each token $\bp_j$, we obtain the cross-attention score as follows:
 $  \mathbf{\hat{m}}_j = \frac{c_j}{\sum_{k} c_k}$,
where $c_j$ is the cross attention score for each token. We proceed by aggregating the attention heads to obtain the average cross-attention scores and resizing them to 32 $\times$ 32, obtaining $\hat{\bm}$. We remove the start of text ($\langle$sot$\rangle$) token for cross-attention and re-normalize it~\cite{chefer2023attend}. In other works~\cite{patashnik2023localizing, ge2023expressive}, as the tokens correspond to the objects being generated in the image, it's sure that token maps will be meaningful. However, this is not true for parts, as Part U-Net might not localize some parts.  For determining if the part is localized, \emph{we propose to look at the max value of cross attention map spatially across the pixels}; if we find that the following condition is met, the part is localized:
\begin{equation}
    L(j) = \indc\{\max (\mathbf{\hat{m}}_j) \geq (1-\delta) \frac{1}{K}\}.
    \label{eq:mask_thresh}
\end{equation}
Here $\delta$ is a hyperparameter, and $K$ is the number of parts. is This condition is robust in finding the localized parts (See Sec.~\ref{sec:analysis} for ablation). For the parts $\bp_i$, which are localized, we normalize the cross-attention map:
\begin{equation*}
    \mathbf{\hat{m}}_{j} =  L(j) \frac{\mathbf{\hat{m}}_{j} - \min(\mathbf{\hat{m}}_{j})}{\max(\mathbf{\hat{m}}_{j}) - \min(\mathbf{\hat{m}}_{j})} + (1 - L(j)) \; \hat{\bm}_j.
\end{equation*}
We follow a dot-product-based protocol to assign each K cluster in the self-attention masks to a part, unlike the average attention protocol in the previous works~\cite{patashnik2023localizing, ge2023expressive}. We find that dot product of normalized cross attention scores $\hat{\bm}_{j}$ for each token with self-attention masks $\mathbf{\hat{M}}_j$ works better, as the attention maps in Part U-Net are noisy. Still, they are often correct for the regions that are localized in one specific area only (Suppl. Fig.~\ref{fig:attention-eval}). Hence, the dot product protocol favors those maps that are only localized in some areas of the image and don't have high attention values across all parts of the object. 
After obtaining the dot product scores for each part, we assign $\hat{\mathbf{M}}_j$ to $\bp_i$ with the highest scores. The mask for the part $\bp_i$ is finally given as the following:

\begin{equation}
    \mathbf{M}_{\bp_i} = \{ \cup_{j} \; \hat{\mathbf{M}}_j|  \argmax_{i} \hat{\mathbf{M}}_j \cdot \mathbf{\hat{m}}_i  = i  \; \text{and}  \; \hat{\mathbf{M}}_j \cdot \mathbf{\hat{m}}_i \geq \epsilon,  \}
    \label{eq:epsilon_thresh}
\end{equation}
Here $\epsilon$ is the hyperparameter which controls the minimum similarity required between the attention mask and the part token. We combine the additional attention masks unassigned to any token, and name them as the background (other) token $\mathbf{M}_{b}$.

\subsection{Part Generation}
We follow a similar protocol for the generation of part segments as in Rich-Text Generation. We tailor the rich text generation to compose the part regions of the object in the image in place of the original scene composition. We describe the part generation protocol below briefly and refer readers to Rich-Text Gen~\cite{ge2023expressive} for more details. For each part $\bp_i$ we run a region diffusion process, which runs in parallel for all the parts. We then combine the region diffusion processes to obtain the final noise prediction $\epsilon_{t}$ as the masked $\mathbf{M}_{\bp_i}$ sum of the denoiser outputs:

\begin{equation}
    \epsilon_{t} = \sum_{i} \mathbf{M}_{\bp_i} \epsilon_{t,\bp_i} = \sum_{i} \mathbf{M}_{\bp_i} \odot D(x_t, f(\bp_i, \ba_i), t)
    \label{eq:part_gen}
\end{equation}
where $D$ is the pre-trained U-Net model, and $f(\bp_i, \ba_i)$ is the text description of the part $\bp_i$ constructed using the following process using the part tokens $\bp_i$ and attributes $\ba_i$. Intially the the text $f(\bp_i, \ba_i) = \bp_i$, is set to part token itself. In case the part description (i.e. footnote) is available we set the $f(\bp_i, \ba_i) = \ba_i^{p}$, further if the style attribute is available we do $f(\bp_i, \ba_i) = f(\bp_i, \ba_i) + \text{in style of} + \ba_{i}^{s}$. In case the color attribute is also specified, the nearest named color $\hat{\ba}^{c}_i$ (\eg red) for the specific RBG color $\ba_{i}^{c}$ is found, and $f(\bp_i, \ba_i) = \hat{\ba}^{c}_i + \text{` '} + f(\bp_i, \ba_i)$. The string $f(\bp_i, \ba_i)$ is the text input for the Diffusion to generate the part $\bp_i$. We use the base prompt $\hat{\mathbf{b}}$ as $f(\bp_i, \ba_i)$ for the background masked region $\mathbf{M}_{b}$. Combining different diffusion outputs at every time $t$ helps generate a harmonious image after blending the defined parts of the object. 
\begin{figure}[!t]
    \includegraphics[width=\columnwidth]{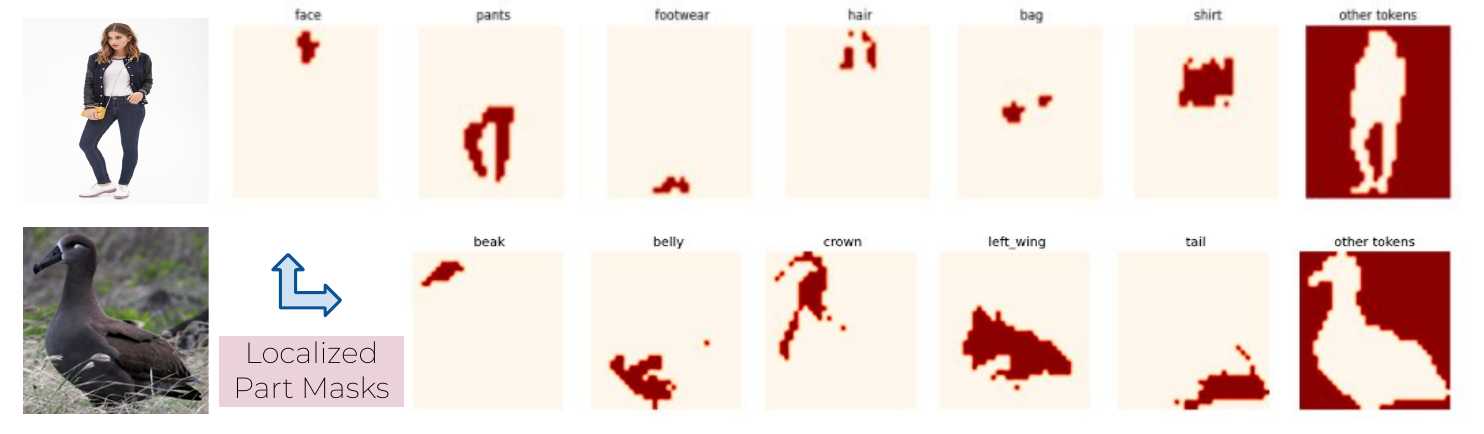}
    \caption{\textbf{Segmentation masks} for the parts that are localized by Part Diffusion for DeepFashion (above) and CUB200 (below).  }
    \label{fig:part-masks-qual}
    \vspace{-5mm}
\end{figure}

Following Rich-Text~\cite{ge2023expressive}, we also utilize the gradient guidance~\cite{ho2022imagen, dhariwal2021diffusion} by taking the gradient of MSE loss between the estimated original image and the color value $\ba_{i}^{c}$ specified by the user. The gradient guidance helps generate the \emph{exact RGB color} for part $\bp_i$, which is impossible with just text guidance~\cite{ge2023expressive}. Further, 
as we use the base text framework of Rich-Text, we also can specify font size attribute to $\ba_{i}^{f}$ to control the relative importance of each part in the object image.

\begin{table*}[!t]
\centering
\caption{\textbf{Comparison to Prior Works} for the unsupervised part segmentation task. We follow Unsup-Part~\cite{choudhury2021unsupervised} for evaluation protocols and baseline results for (K=4) parts. We report clustering-based NMI and ARI metrics, which are higher for better segmentation outputs. }
\vspace{-5pt}
\begin{tabular}{lcccccccc}
\toprule
\multicolumn{5}{c}{DeepFashion Dataset} & \multicolumn{4}{c}{CUB200 Dataset} \\
\cmidrule(r){1-5} \cmidrule(l){6-9}
Method & FG-NMI & FG-ARI & NMI & ARI & FG-NMI & FG-ARI & NMI & ARI \\
\midrule
& \multicolumn{8}{c}{Unsupervised Learning} \\
\midrule
SCOPS~\cite{hung2019scops} & 30.7 & 27.6 & 56.6 & 81.4 & 39.1 & 17.9 & 24.4 & 7.1 \\
Unsup-Parts~\cite{choudhury2021unsupervised} & 44.8 & 46.6 & 68.1 & 90.6 & 46.0 & 21.0 & 43.5 & 19.6 \\
DFF~\cite{choi2019self} & {-} & {-} & {-} & {-} & 32.4 & 14.3 & 25.9 & 12.4 \\
\midrule
& \multicolumn{8}{c}{\textbf{Unsupervised Zero-Shot}} \\
\midrule
Rich-Text~\cite{ge2023expressive}  & 16.0 & 5.2 & \textbf{48.3} & 58.7 & 3.1 & 0.3 & 3.1 & 0.3\\
StableDiffusion & 12.0 & 3.5 & 40.6 & 70.9 & 8.0 & 0.6 & 3.3 & 0.6 \\
PartComposer(Ours) & \textbf{24.7} & \textbf{18.0} & \textbf{48.0} & \textbf{73.4} & \textbf{20.5} & \textbf{9.2} & \textbf{18.5} & \textbf{7.7} \\
\bottomrule

\end{tabular}

\vspace{-5mm}
\label{tab:metrics-part-loc}
\end{table*}

\noindent \textbf{Preservation of Other Parts.} As we only intend to modify the object $\mathbf{o}$ from the original prompt, we also use the Self-Injection techniques from Plug and Play~\cite{tumanyan2023plug} to maintain the overall structure of the background from base prompt generation. Further, to ensure that our diffusion trajectory follows the same path as the base, the background region is also blended with base noise generation outputs. 
\begin{equation}
    x_t = \mathbf{M}_{b} \odot x_{t}^{\text{base}} + (1 - \mathbf{M}_{b}) \odot x_t
    \label{eq:self_attn_inj}
\end{equation}
In our case, the attention maps out of the object Mask $\mathcal{M}_{o}$ and all the object parts not assigned to any part token comprise the background $\mathbf{M}_{b}$, and we start this blending process at $t = T_{blend}$. We find this to be very useful in preserving the structure of the other parts of the image and just generating the described parts in the localized region. We provide an overview of the complete process in Fig.~\ref{fig:overview}.

%% file: paper-sections/experimental-analysis.tex
\subsection{Evaluation of Part Localization}

We first evaluate the part localization module, which is based on the novel idea of denoising only the object region with the part-based diffusion outputs.

\vspace{1mm} \noindent \textbf{Implementation Details}. We use the StableDiffusion (SD) version 2.1 for our experimentation purposes. We use the DDIM Scheduler~\cite{song2021ddim} with 50 steps to generate results for SD2.1 to evaluate the part localization process. As ground truth is not available for the part masks of the generated images, we use test sets of the commonly used DeepFashion~\cite{liuLQWTcvpr16DeepFashion} and CUB-200~\cite{WahCUB_200_2011} datasets for evaluation. These datasets are the standard datasets used for the evaluation of the unsupervised part segmentation approaches. The DeepFashion dataset contains images along with their part segmentation masks, divided into 14 categories of labels. The CUB200 dataset contains the key point annotations for the 14 key points specified for the bird categories. We provide further details in Suppl. Section.

\vspace{1mm} \noindent \textbf{Baselines and Problem Setting. } As we operate in the setting of Zero-shot (\ie no training) Unsupervised Part Segmentation, there are no previous works that report results in such a challenging setting. Hence, we provide results for the unsupervised learning approaches to facilitate comparison. We provide results for the SCOPS~\cite{hung2019scops}, which utilizes the internal features of the VGG model to train a model based on self-supervised loss functions to predict parts robustly across categories. 
The other stronger self-supervised baseline is Unsup-Parts~\cite{choudhury2021unsupervised}, which uses contrastive loss functions to train a network based on equivariance and other vision properties to cluster the object regions into semantic parts. 
In addition to this, we also report results for the DFF~\cite{choi2019self} as they also operate in the same setting. We want to highlight that these unsupervised approaches \emph{require either training a neural network or performing clustering on the complete training data to segment parts}. In contrast, our approach is \emph{training free and operates in a zero-shot fashion}. Hence, the performance of zero-shot approaches is not directly comparable to unsupervised methods.

\noindent \textbf{Zero Shot Unsupervised Part Segmentation. } As there is no benchmark to evaluate the part localization for generated images, we use the existing dataset of DeepFashion and CUB200 to obtain our results. To obtain segmentation for each image, we first invert the image into the diffusion latent space using Null-Text Inversion~\cite{mokady2023null} method (see Sec.~\ref{sec:analysis} for ablation). We use the BLIP-V2~\cite{li2023blip} captioner provided in Diffusers to obtain approximate text prompts for inversion. After providing the desired image and prompt, the Null-Text inversion provides us with inverted latent and unconditioned time embeddings to reconstruct the image of the dataset. We then construct a StableDiffusion baseline in which, in addition to the prompts, we append the list of part tokens $\bp$ to the prompt. We then use the segmentation algorithm described above in Sec.~\ref{subsec:part_loc} to extract token maps corresponding to the part tokens $\bp_i$. We also evaluate a Rich-Text~\cite{ge2023expressive} baseline segmentation algorithm for the same. For the proposed part-denoising approach based on parallel diffusion in PartComposer, we first get the inverted latents and embeddings. Then, we generate a base diffusion process to reconstruct the image and use Part Diffusion to fill the parts of the image. To very fairly compare the StableDiffusion performance with part-denoising, we keep all things the same except the part diffusion process to get masks. 
Further across these baselines, we use low classifier guidance to ensure proper reconstruction of the dataset images (See Suppl. for details).

\begin{figure*}[!t]
    \centering
        \includegraphics[width=0.95\textwidth, height=5.5cm]{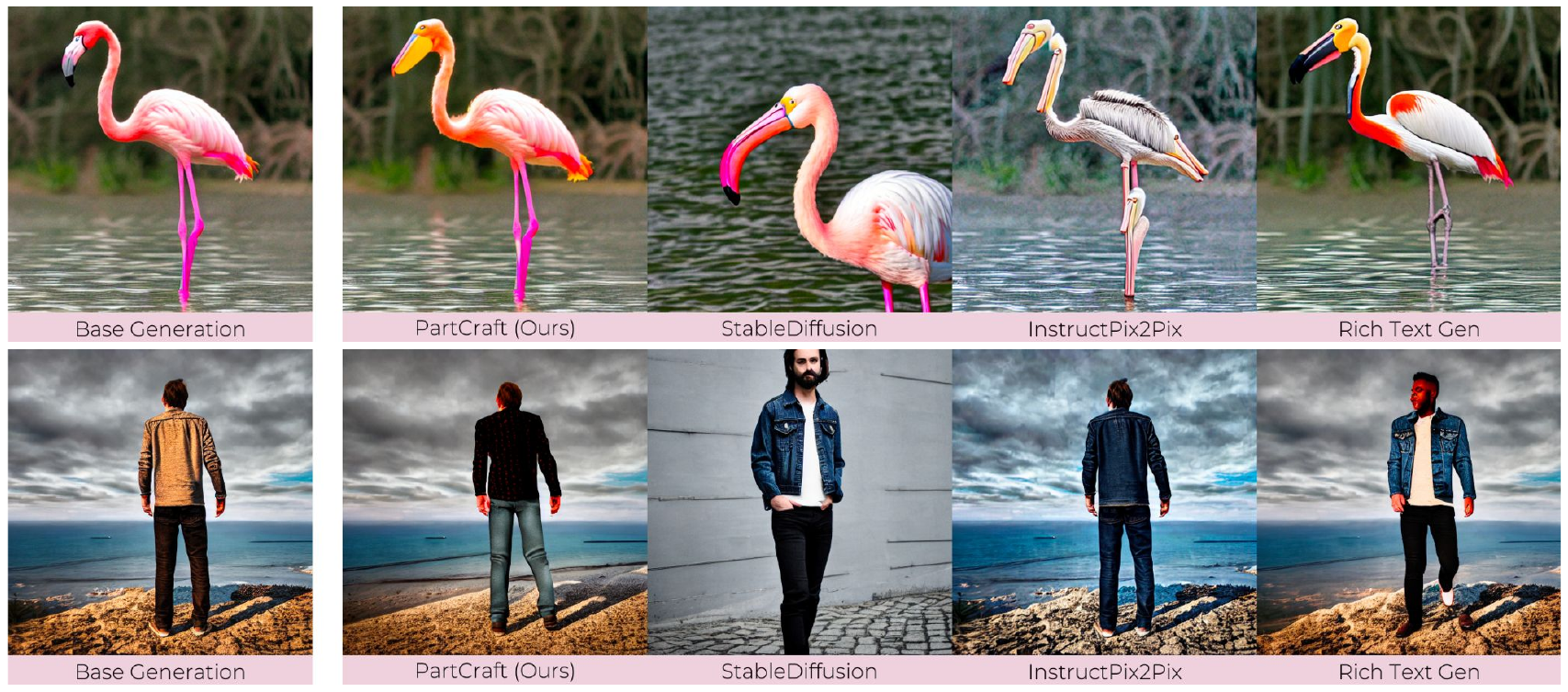}
    \vspace{-2mm}
    \caption{\textbf{Qualitative Comparison for PartComposer.} (above) We provide base prompt `a photo of a flamingo' and part prompt as `beak - a pelicans beak'. (below) We generate a photo of a man, with part prompt as `black jacket and blue jeans'. We find that other baselines either ignore instruction or change the entire composition. On the contrary, PartComposer can correctly localize parts and generate details.}
    \label{fig:qual-comparison-birds}
    \vspace{-5mm}
\end{figure*}

\noindent \textbf{Evaluation Protocol.} We use the standard experimental protocol as used by earlier works~\cite{choudhury2021unsupervised} to facilitate the right point of comparison. We want to point out that in our work, the part name (\eg, beak, etc.) is associated with the localized mask, but the unsupervised approaches produce (K=4) parts without any part names. Hence, to make a fair comparison, we create 4 clusters of parts based on their locality, and finally generate segmentation masks with a maximum of 4 clusters. We provide the exact mapping of the part names to cluster labels in the Suppl. Section. We compare the specifically designed metrics~\cite{choudhury2021unsupervised} of NMI (Normalized Mutual Information) and ARI (Adjusted Rand Index) of the predicted cluster labels with part masks in the case of DeepFashion and key points in case of the CUB dataset. We report the metrics along with their foreground variants (FG-NMI and FG-ARI) for a holistic comparison of performance.

\noindent \textbf{Results.} We tabulate the results for all the baselines in Table~\ref{tab:metrics-part-loc}. Our approach, PartComposer, significantly improves over the baselines by $\geq$ 5 points in NMI and ARI across most metrics in the Unsupervised Zero-Shot setting. Further, in some cases our methods, NMI and ARI, are near the unsupervised learning approaches. This demonstrates that non-trivial part masks can be generated by harnessing the power of the pre-trained diffusion models, and they can serve as a good initial prior for learning part segmentation approaches by using them as base pre-trained models. We provide qualitative results for PartComposer, where we observe that PartComposer can associate the right part with the correct label (Fig~\ref{fig:part-masks-qual}). On the contrary, in the StableDiffusion baseline, the parts often get assigned to the wrong part tokens, which is the cause of degraded performance (Suppl. Fig.~\ref{fig:qual-masks-cub}). We also observe that in the case of our algorithm, if the part gets localized based on the max condition defined in the segmentation algorithm, the part mask found is usually in the right region (See Fig.~\ref{fig:part-masks-qual} and Table~\ref{tab:partcraft-ablation-main}). This demonstrates the effectiveness of our part segmentation procedure, which often sides with not localizing objects rather than performing arbitrary assignments.

\subsection{PartComposer Image Generation Evaluation}
We now evaluate the object composition ability of the proposed approach, PartComposer, compared to the baselines. In this section, we use the base generation model as a StableDiffusion (SD) 1.5 model to do a fair comparison with the Rich-Text baseline. We used the same generation setting of a PNDM~\cite{liu2022pseudo} scheduler with 41 steps and suggested classifier guidance 8.5. To perform a fair comparison, we keep all the parameters same as that of the Rich-Text~\cite{ge2023expressive} baseline. 

\noindent \textbf{Baselines.}
For the task of generating the object image based on part-level details of the specified object, we use the strong baselines of Rich-Text Generation and InstructPix2Pix~\cite{brooks2022instructpix2pix}. The Rich-Text Generation method generates the base image itself, whereas the InstructPix2Pix method requires us to generate the base image. We also evaluate the standard StableDiffusion baseline in which we add all the part details in the text prompt to generate the desired image. We compare all these baselines with our proposed method, PartComposer, while ensuring that the base parameters and seed are the same for the base image generation. A recent work~\cite{ng2024partcraft} regarding part generation operates in a fully supervised setup using part masks and cannot be used to create novel parts based on text instruction, making it incomparable. We defer specific details for the baselines to the Suppl. Section. 

\noindent \textbf{Visual Comparison.} We provide a visual comparison for the baselines for the a) bird image distinctive part generation and b) human image part generations. Across both cases (Fig.~\ref{fig:qual-comparison-birds}), our proposed method, PartComposer, only modifies the desired beak region and replaces it with the iconic pelican beak. The StableDiffusion baseline modifies the image completely while ignoring instructions (Supp. Fig.~\ref{fig:add-qual-results}). In the Rich-Text generation, the full region corresponding to the bird gets modified instead of only the specified prompt in base generation. For the editing-based InstructPix2Pix method, we observe that it modifies the base image at a global level and cannot localize modification to the desired part region (Fig.~\ref{fig:qual-comparison-birds}). We further compare and find that recent models like SD3.5, Inpainting~\cite{esser2024scaling} and recent SotA editing methods~\cite{deutch2024turboedittextbasedimageediting, tumanyan2023plug, zhang2023magicbrush, li2024zone} can still not follow exact part instructions as seen in Supp. Fig.~\ref{fig:sd3.5_inpaint},~\ref{fig:edit-comp} and ~\ref{fig: sd-xl-full}.
In contrast,observe that PartComposer-generated images are novel aesthetic compositions that follow the part-level instructions. 

\begin{figure}
  \caption{\textbf{User Study and Quantitative Results} for part-based image generation baselines.}
     \label{fig:user-eval}
     \vspace{-4mm}
     \centering
    \includegraphics[width=0.49\textwidth]{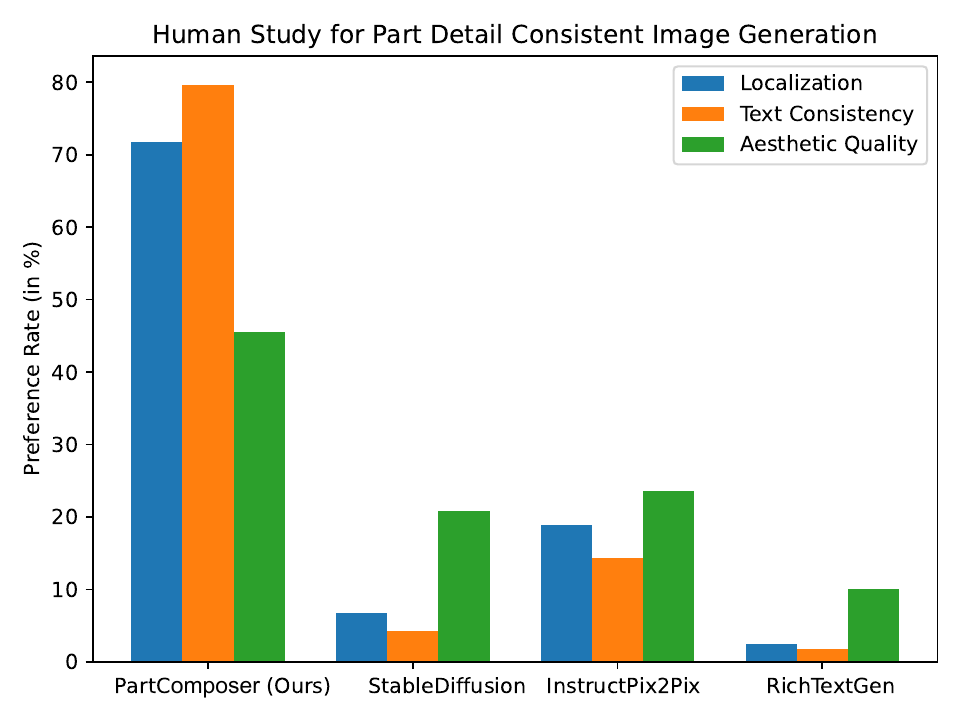}
    
  \adjustbox{max width=\columnwidth}{%
    \centering
    
    \begin{tabular}{l||c|c|c|c}
    
          &PartComposer & StableDiffusion  & InstructPix2Pix &  Rich-Text \\ \hline
        LPIPS & \textbf{0.168} & 0.467 & 0.189 &  0.243 \\ 
        CLIP & \textbf{0.201} & 0.183 & 0.193 &  0.187 \\ 
        Aesthetic & 5.66 & 5.68 & 5.63 &  5.65 \\ \hline
    \end{tabular}}
  
  \vspace{-6mm}
\end{figure}

\noindent \textbf{Quantitative Evaluation.} As the part-specific generation is fine-grained, we perform the both automatic and user-study evaluations. We performed a user study by inviting 50 participants (28 responded with a full survey). 
We evaluate the model by asking questions and evaluating metrics on three orthogonal aspects i.e., a) \textbf{Localization} of part generation in comparison to the base image (through LPIPS~\cite{zhang2018unreasonable}), b) \textbf{Text-Consistency} of generated parts by measuring CLIP similarity of the part detailed text prompt with the generated image and c) \textbf{Aesthetic Quality} of the generated image through LAION-5B~\cite{schuhmann2022laion} CLIP Aesthetic scorer. 
In total, we collect about 3.5k opinions. In Fig.~\ref{fig:user-eval}, we summarize the results in which we observe that PartComposer is significantly preferred over the other baselines in terms of localized and consistent Part Generation. 
The automated evaluation results are provided in the table below, which also follow a similar trend as in the user study, demonstrating the effectiveness of PartComposer in a localized part generation while ensuring the aesthetics score is similar to that of the base model.

\begin{table}[!t]
    \vspace{-2mm}
  \caption{\textbf{Ablation Analysis} of PartComposer for Part Localization, showing performance improvement with each component.}
 \vspace{-2mm}
  \label{tab:partcraft-ablation-main}
\begin{adjustbox}{max width=\textwidth}
    \begin{tabular}{lcc}
      \toprule
      \multicolumn{3}{c}{\textbf{PartComposer (Ablations)}} \\
      \midrule
      Method & FG-NMI & FG-ARI \\
      \midrule
      PartComposer & 35.4 & 11.0 \\
      w/o Null-Text Inversion & 23.1 & 5.2 \\
      w/o Max Localization & 21.3 & 2.8 \\
      w/o Dot Product Localization & 23.7 & 5.0 \\
      w/o Independent Text & 31.2 & 8.9 \\
      
      \midrule
      \multicolumn{3}{c}{\textbf{PartComposer (Clustering)}} \\
      \midrule
      PartComposer (K = 9) & 35.4 & 11.0 \\
      PartComposer (K = 4) & 20.4 & 2.8 \\
      PartComposer (K = 14) & 35.2 & 10.3 \\
      \bottomrule
       \vspace{-5mm}
    \end{tabular}
    
  \end{adjustbox}

    \label{tab:table2}
  \end{table}

\begin{figure}[!t]
    \centering
        \includegraphics[width=\columnwidth]{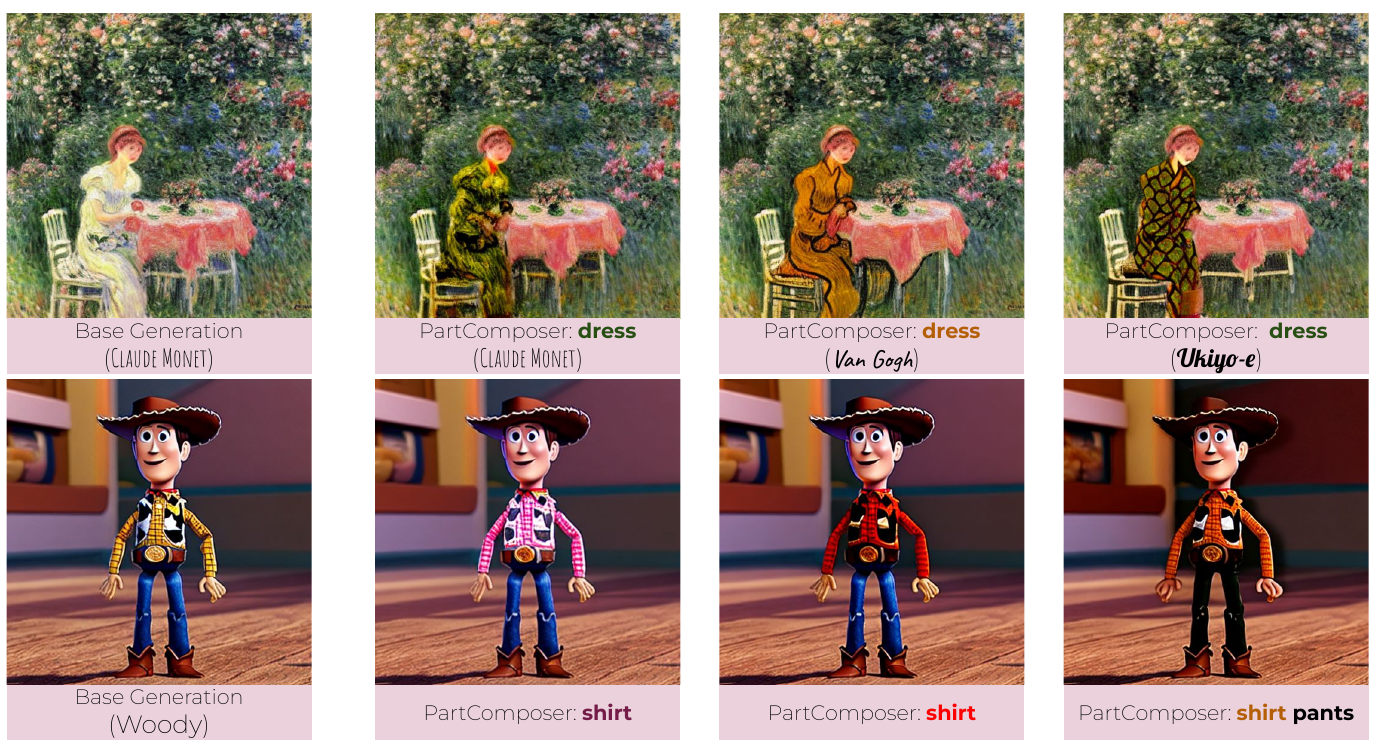}
    \caption{\textbf{PartComposer generalizes over domains} as shown by Claude Monet's painting style (above), where we specify the dress part of women to follow styles like Van Gogh and Ukiyo-e. We perform similar modifications to Woodie's image.    }
    \vspace{-5mm}
    \label{fig:generalization-domains}
\end{figure}

\section{Analysis and Discussion}
\label{sec:analysis}

\noindent \textbf{Part Diffusion Based Localization.} 
We ablate the components we have introduced in the PartDiffusion process of PartComposer (Sec.~\textcolor{red}{3.2}). We provide an analysis of the effect of using \textbf{a)} null-text inversion, \textbf{b)} max-based localization \textbf{c)} usage of dot product-based protocol in the part assignment, and \textbf{d)} independent text embeddings. We have used a subset of CUB-200 images to perform all evaluations, which is kept fixed across ablations. We tabulate the ablations in Table~\ref{tab:partcraft-ablation-main}. We observe that all the components introduced in PartComposer contribute significantly to the performance.

\noindent \textbf{Generalization of PartComposer.}  In Fig.~\ref{fig:generalization-domains}, above a painting in Claude Monet style generated by the base StableDiffusion method. We then use PartComposer to specify the \textbf{dress} part of the women's token, to dark green color in Monet Style, orange color dress in Van Gogh Style (middle), and green color dress in Ukiyo-e style.  In Fig.~\ref{fig:generalization-domains}, we provide results for the `Woody' character from Toy Story. These results show that PartComposer can generate shirt, pants, and dress color variations, leading to aesthetic image combinations in synthetic and natural (Supp. Fig.~\ref{fig:att_variants}) domains . 
This zero-shot generalization across domains demonstrates the creative activities that can be enabled with PartComposer.

%% file: paper-sections/conclusion.tex
In this work we introduce PartComposer, a method to generate object images with fine-grained attribute details specified at the part level, using an expressive Rich-Text interface. The PartComposer method introduces a novel part diffusion process, which is responsible for denoising objects using the part features, and then utilizes a region-specific diffusion process to generate part details and compose the final image. PartComposer serves as initial work enabling rich-text-based training-free part-level control for SD models. \\
\noindent \textbf{Limitations.} We find that part generation is bottlenecked with the part understanding. If the part is localized correctly, the text-to-image model can generate specified details in PartComposer. Hence, improving the part-level localization of these models is a good direction for future works.

%% file: paper-sections/supp.tex
\maketitlesupplementary
\setcounter{page}{1}

\appendix

\addcontentsline{toc}{section}{} %

\part{} %
\parttoc %

\section{Additional Qualitative Results}

\begin{figure*}[!t]
    \centering
    \includegraphics[width=\textwidth]{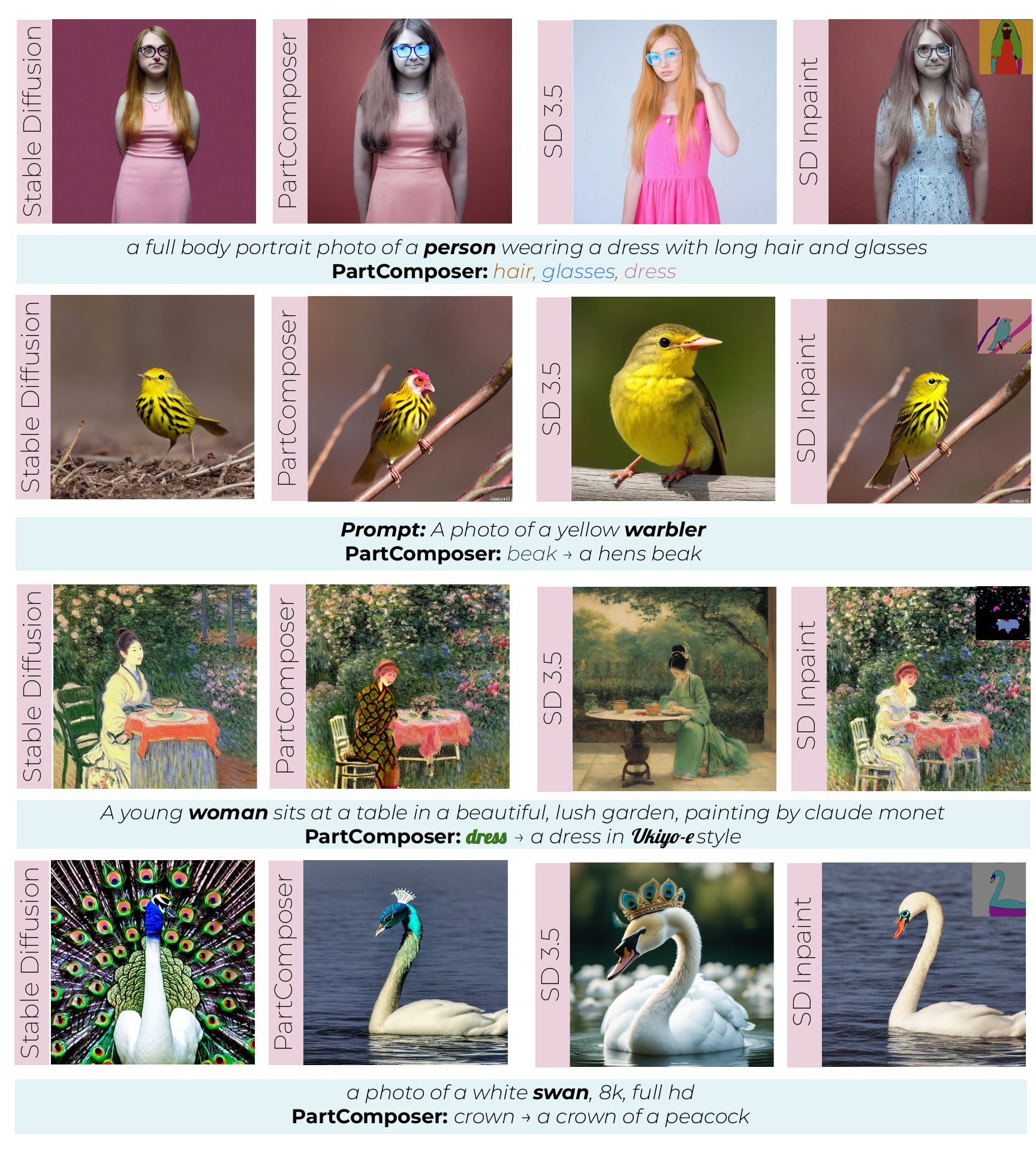}
    \caption{\textbf{Qualitative Comparison} of our proposed method with recent baselines of Stable Diffusion (SD) 3.5 and Stable Diffusion 2.1 Inpainting with SAM2 masks (top right). PartComposer leads to coherent part modification in comparison to advanced baselines.}
    \label{fig:sd3.5_inpaint}
\end{figure*}

\vspace{1mm} \noindent \textbf{Comparison with Recent Baselines.} We compare PartComposer against existing baselines of StableDiffusion 3.5 ~\cite{esser2024scaling} and Inpainting through Stable Diffusion. For the StableDiffusion (SD) 3.5 model, we use the SD3.5 Large variant for which the web-UI is available online at the following link\footnote{\href{https://huggingface.co/spaces/stabilityai/stable-diffusion-3.5-large}{https://huggingface.co/spaces/stabilityai/stable-diffusion-3.5-large}}. We provide the details of the PartComposer by adding those meaningfully into the base prompt. For the swan example in the last row of Fig.~\ref{fig:sd3.5_inpaint}, we provide the following prompt to the model \emph{`A photo of a white swan with a peacock crown, 8k, full hd'}. We observe in Fig.~\ref{fig:sd3.5_inpaint} that although the model tries to adhere to the prompt in ways, it fails to modify the correct object part. Hence, it fails to capture the required details of the part the user expects. For the Inpainting baseline, we use the same StableDiffusion (SD) 2.1 model as a base and use the SAM2 Hiera~\cite{ravi2024sam} model for obtaining the segmentation masks. We use the following inpainting repo\footnote{\href{https://github.com/Uminosachi/inpaint-anything}{https://github.com/Uminosachi/inpaint-anything}} from GitHub to obtain the results. After obtaining the masks, we provide an inpainting prompt to the SD model to generate the final result. We observe that the SD Inpaint models sometimes, due to incorrect masks for the painting domain, don't produce results (third row in Fig.~\ref{fig:sd3.5_inpaint}) and often can't follow the multiple-part instruction correctly as in Fig.~\ref{fig:sd3.5_inpaint} (first row). Further we would like to highlight that inpainting still requires manual ground of masks to parts by user, which makes it a different setting making the quantitative results incomparable. In comparison, PartComposer can produce correct and coherent results by producing the part attributes.

\begin{figure*}[!t]
  \centering
  \includegraphics[width=\linewidth]{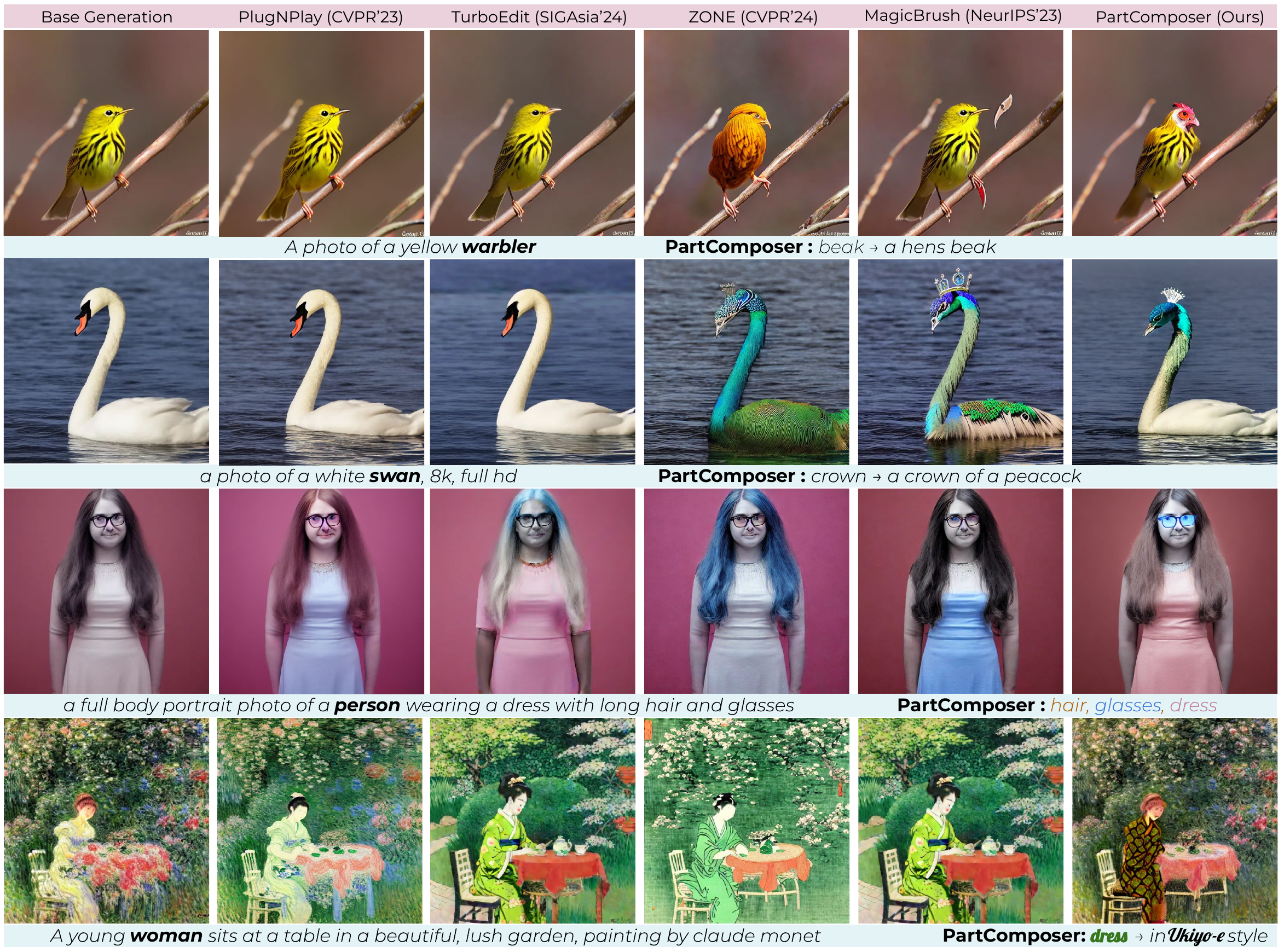}
   \caption{\textbf{Qualitative Comparison} of PartComposer with SotA editing methods applied on the generated base prompt image.}
   \label{fig:edit-comp}
\end{figure*}

\vspace{1mm} \noindent \textbf{Comparison with Editing Methods.}
We provide a comparison of our approach with SotA editing methods:   PlugNPlay (PnP)~\cite{tumanyan2023plug}, TurboEdit~\cite{deutch2024turboedittextbasedimageediting}, ZONE~\cite{li2024zone}, MagicBrush~\cite{zhang2023magicbrush} in Fig.~\ref{fig:edit-comp}.  We observe that approaches either do style edits (PnP) or
structure edits at full object level (ZONE, MagicBrush). \ul{Only PartComposer (ours) can perform the specified localized changes at the semantic part level, demonstrating its novelty and value.}

\begin{figure*}[!h]
    \centering
    \includegraphics[width=\textwidth]{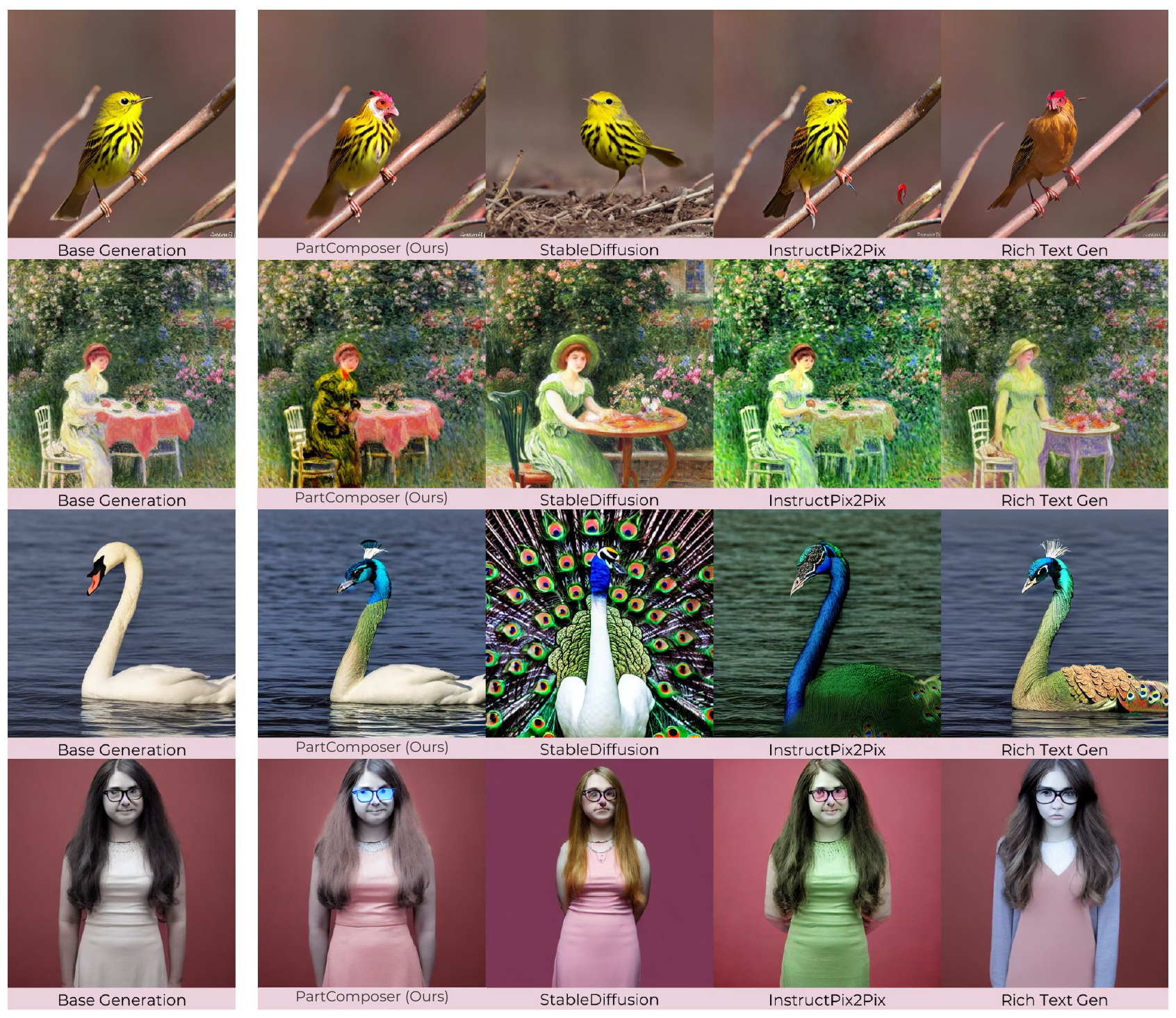}
    \caption{\textbf{Comparison of PartComposer to Other Approaches}, for generating using the prompts specified in Fig.~\ref{fig:teaser_qual} of paper. The PartComposer approach can modify base generations in the specified parts and generate aesthetic images compared to the state-of-the-art.}
    \label{fig:part-craft-qualitative-results}
\end{figure*}

\vspace{1mm} \noindent \textbf{Additional Comparison of PartComposer to Baselines (Fig.~\ref{fig:teaser_qual}).} In Fig.~\ref{fig:part-craft-qualitative-results}, we demonstrate a comparison of the StableDiffusion, InstructPix2Pix, and Rich-Text on the prompts demonstrated in the teaser figure. For Rich-Text, we add the description in PartText to the part token if it's present in the base prompt; if not, we add the part description as a footnote in the object token in the base prompt. For InstructPix2Pix, we make instructions regarding part modification one by one on the base generation. We find that the baselines significantly change the entire composition of the object instead of just specified parts. 

\begin{figure*}[!h]
    \centering
    \includegraphics[width=\textwidth]{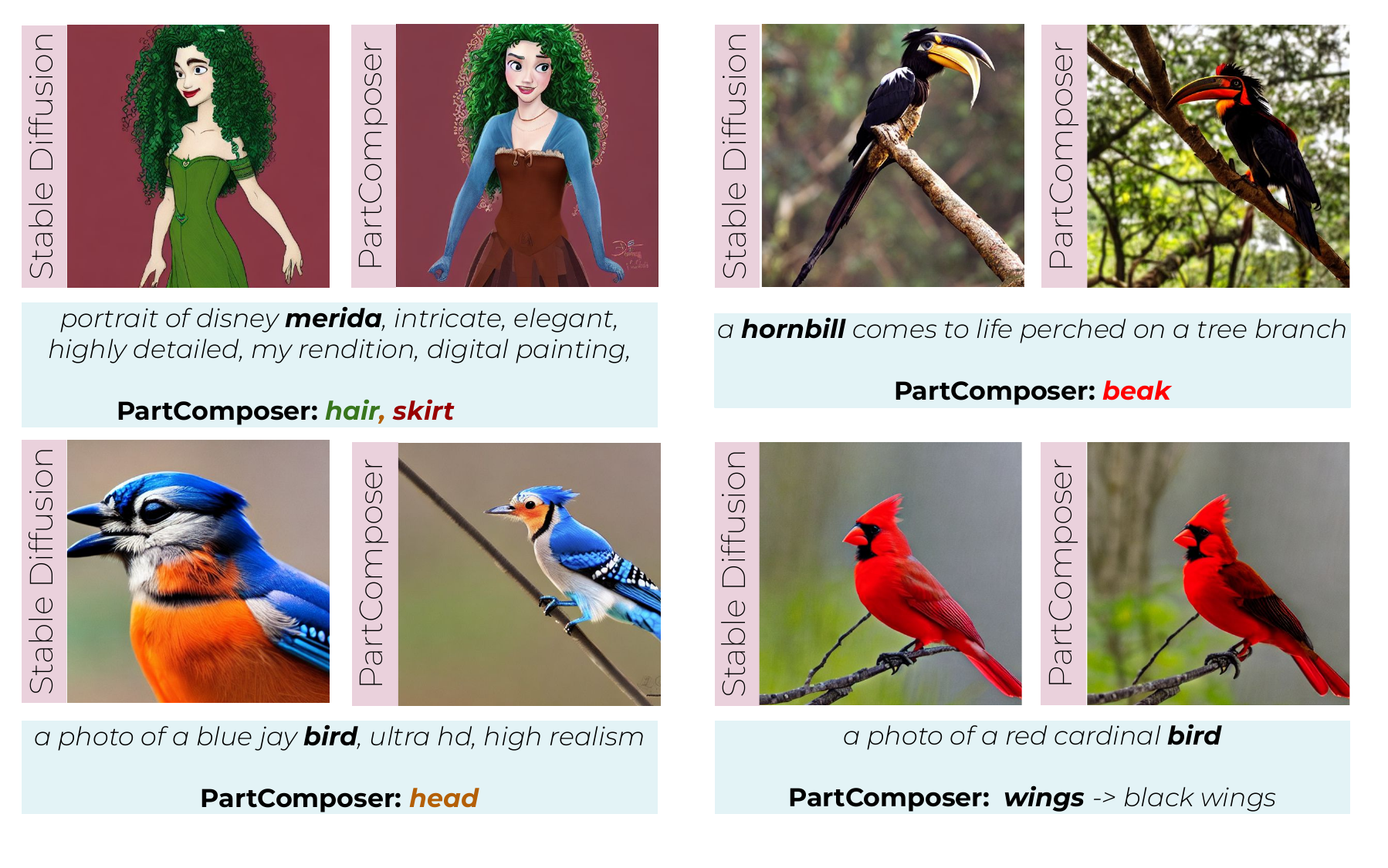}
    \caption{\textbf{Qualitative Results.} We provide a qualitative comparison of additional prompts with Stable Diffusion. We add the part attribute details to the prompt during generation for Stable Diffusion baseline. The PartComposer can generate the image following the specified part attribute details with better coherence and fewer artifacts compared to the baseline.}
    \label{fig:add-qual-results}
\end{figure*}

We also provide additional results for comparison with baselines in Fig.~\ref{fig:add-qual-results}, where the Stable Diffusion baseline with part details added often ignores them or generates artifacts. In comparison, PartComposer can produce the desired details for the parts as specified by the user. As in Merida's example, the Stable Diffusion method doesn't produce the brown skirt and over uses the green color specified for hair. The Stable Diffusion ignores the desired part detail for the hornbill and cardinal examples. The blue jay example produces an artifact of the dual beak. In comparison, PartComposer generates aesthetic compositions following the part prompt details mentioned by the user. 

\section{Generalization to SDXL}
\begin{figure*}[!t]
  \centering
  \includegraphics[width=\linewidth]{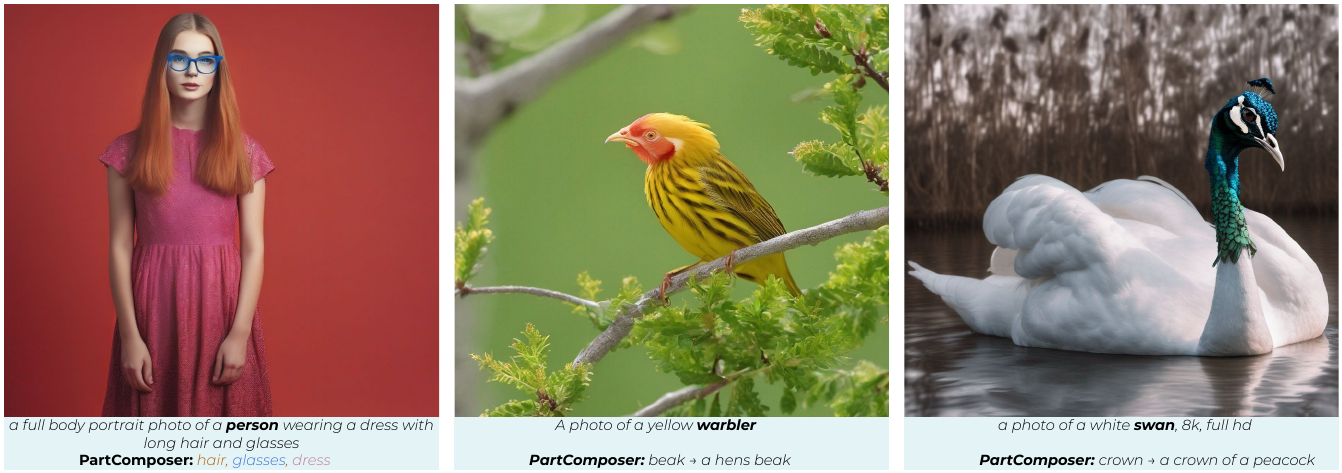}
   \caption{\textbf{High-Resolution} PartComposer results using SD-XL as base model, demonstrating its generalization.}
   \label{fig: sd-xl-full}
\end{figure*}

We implement the PartComposer Method for SDXL, and generate the results for prompts supplied in Fig.~\ref{fig:teaser_qual}. Using the SD-XL implementation from Rich-Text Gen~\cite{ge2023expressive} as our base code, we implement PartComposer. We present our results in Fig.~\ref{fig: sd-xl-full}, where we observe that PartComposer can generate high-resolution images with specific attribute details.

\section{Attribute Variations}
\begin{figure}[!t]
    \centering
    \includegraphics[width=\columnwidth]{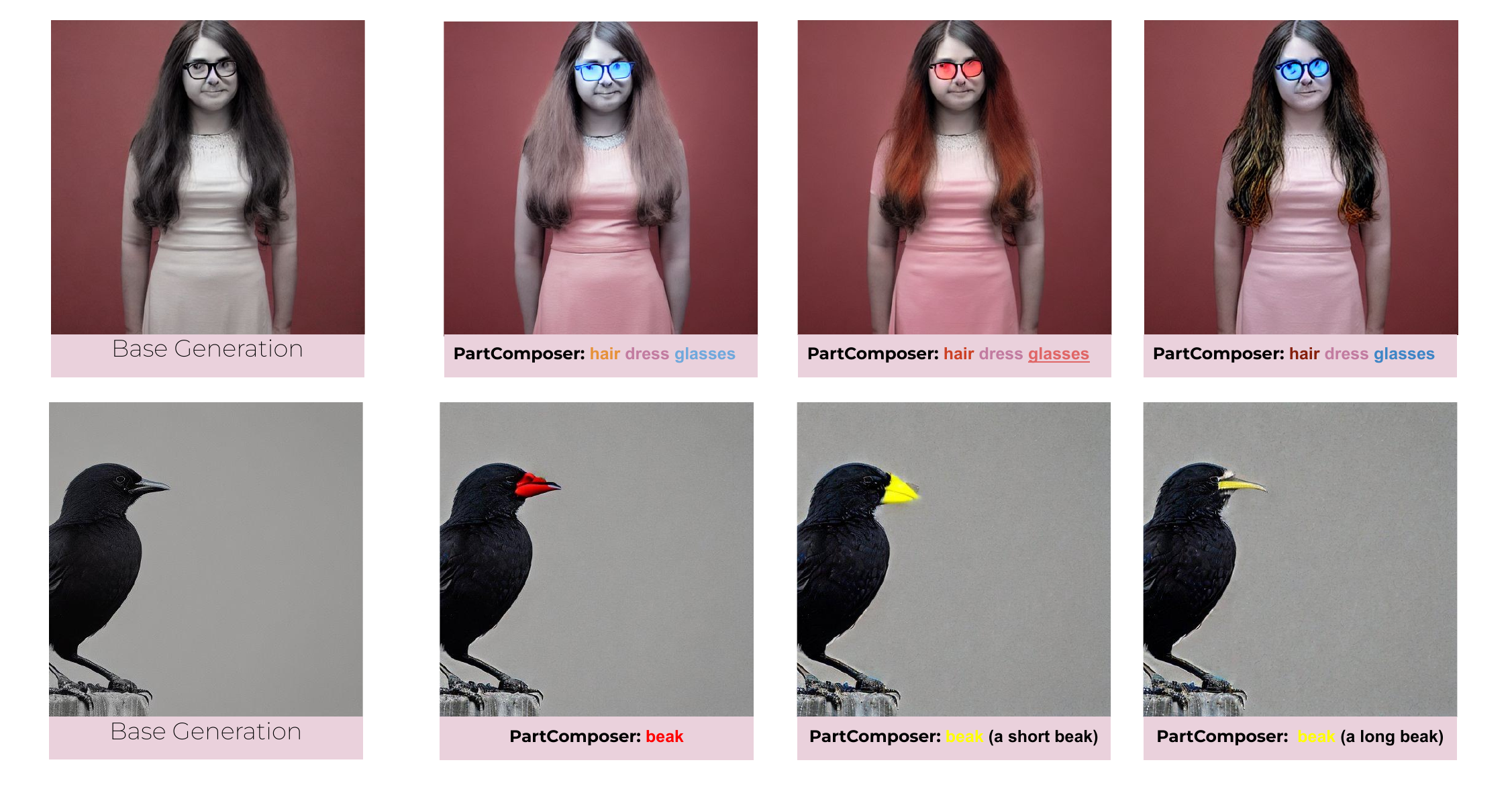}
    \caption{\textbf{Attribute Variations} for the base generated images are provided for the person portrait, where we change color of hair and eyeglasses, increase weight of eye glasses to change its shape. In the bird example, we generate various colors of beaks and generate its part variations by specifying part description in bracket.}
    \label{fig:att_variants}
    
\end{figure}

We provide further results for attribute variations for prompts of natural domain in Fig.~\ref{fig:att_variants}.

\null\newpage
\null \newpage

\null\newpage
\null \newpage

\null\newpage

\null\newpage
\null \newpage

\section{Notations}
We provide notations used in the paper in Table~\ref{tab:notations}.

\begin{table}[]
\centering
\caption{\textbf{Notation Table} for the paper.}
\label{tab:notations}
\resizebox{\linewidth}{!}{%
\begin{tabular}{p{0.1\linewidth} p{0.65\linewidth}}
\toprule
Symbol                & Meaning                                          \\ \midrule
$\mathcal{M}_{o}$ & Mask of Object $\mathbf{o}$ \\
$\hat{\mathbf{b}}, \hat{b}$ & Set of Base Prompt Tokens \\
$\hat{\bp}$ & Set of Part Prompt Tokens \\
$T_{th}$ & Threshold After Which The Part Diffusion Starts \\
$\hat{\bM}_j$ & Self Attention Map ($j^{th}$ index) \\
$\hat{\bm}_k$ & Cross Attention Maps (for k$^{th}$ token) \\
$x \cdot y$ & Dot Product Between x and y, viewed as vectors \\
$A \odot B$ & Elementwise (Hadamard) product between A and B \\
$\mathbf{M}_{\bp_i}$ & Part Mask for the part $\bp_i$ \\
$\ba_i$ & Attribute Description of part $\bp_{i}$ \\
$f(\bp_i, \ba_{i})$ & Text Description of part $\bp_i$ with attributes $\ba_i$ \\
$x_t$ & Partial Denoised Image at time $t$ \\
$\epsilon_{t}$ & Noise output for estimated noise at time $t$ \\
$D$ & Denoising U-Net \\
$\bM_b$ & Background (Other) Token Mask \\
$\alpha$ & Hyperparameter for the Part Diffusion Contribution \\
$\delta$ & Hyperparameter for Part Selection \\
$\epsilon$ & Hyperparameter for the Object Categorization to Background \\

\bottomrule
\end{tabular}}

\end{table}

\section{Potential Negative Impact}
Text-to-image generative models have shown great promise in image generation and can be utilized in content and media creation. However, ensuring the created content is unbiased, harmless, and free of misinformation is essential. Our work gives users more control over text-to-image generation, which should be used responsibly to avoid misinformation. 

\section{Implementation Details}
We discuss implementing the two parts of the PartComposer process, the Part Localization and Part Generation steps. 

\vspace{1mm} \noindent \textbf{Part Localization.} In Part Localization, we first discuss the implementation on StableDiffusion 2.1, where we use 50 steps of DDIM Scheduler~\cite{song2021ddim} for denoising. As we aim to evaluate our method on CUB-200 and DeepFashion datasets, we first perform inversion using Null-Text Inversion~\cite{mokady2023null} with a guidance scale of 0.05 (other hyper-parameters are kept default). We keep the $\alpha$ value linearly scaled with time steps from 0 to 0.5 (Eq.~\ref{eq:alpha-mask}). To obtain the mask for each segment, we keep a low $\epsilon = 0.05$ threshold (Eq.~\ref{eq:epsilon_thresh}), as here, we want to segment the region in all possible parts as the DeepFashion and CUB-200 generate localization masks for the 14 part regions. However, for comparison with baselines producing segments in 4 parts, we perform the following clustering of parts in the four cluster regions (Table~\ref{tab:cub_deepfashion}). We use (K=9) to cluster the different regions into 4 clusters. We use the foreground masks for objects provided by Choudhury \etal~\cite{choudhury2021unsupervised}\footnote{https://github.com/subhc/unsup-parts} for evaluating the FG-NMI and FG-ARI metrics in Table~\ref{tab:cub_deepfashion}.

\begin{table*}[!t]
\centering
\caption{\textbf{Part Names} in Clusters for CUB and DeepFashion Datasets}
\label{tab:cub_deepfashion}
\begin{adjustbox}{max width=\textwidth}
\begin{tabular}{|c|l|l|}
\hline
\textbf{Cluster} & \textbf{CUB Part Names} & \textbf{DeepFashion Part Names} \\ \hline
0                & background               & background                       \\ \hline
1                & beak, forehead, left eye, right eye & cap, hair \\ \hline
2                & breast, crown, nape, throat & dress, shirt (top) , accessories, outer                \\ \hline
3                & belly, left leg, right leg, tail & glasses, face, body          \\ \hline
4                & back, left wing, right wing &                  pants, footwear, leggings             \\ \hline
\end{tabular}
\end{adjustbox}
\end{table*}
\begin{figure*}[!t]

\begin{minipage}[!t]{0.49\textwidth}
   
  \captionof{table}{\textbf{Ablation Analysis} of PartComposer for Part Localization.}
 \vspace{2mm}
  \label{tab:partcraft-ablation}
  \centering
\begin{adjustbox}{max width=\textwidth}

    \begin{tabular}{lcc}
      \toprule
      \multicolumn{3}{c}{\textbf{PartComposer (Ablations)}} \\
      \midrule
      Method & FG-NMI & FG-ARI \\
      \midrule
      PartComposer & 35.4 & 11.0 \\
      w/o Null-Text Inversion & 23.1 & 5.2 \\
      w/o Max Localization & 21.3 & 2.8 \\
      w/o Dot Product Localization & 23.7 & 5.0 \\
      w/o Independent Text & 31.2 & 8.9 \\
      
      \midrule
      \multicolumn{3}{c}{\textbf{PartComposer (Clustering)}} \\
      \midrule
      PartComposer (K = 9) & 35.4 & 11.0 \\
      PartComposer (K = 4) & 20.4 & 2.8 \\
      PartComposer (K = 14) & 35.2 & 10.3 \\
      \bottomrule
    \end{tabular}
  \end{adjustbox}

    \label{tab:table2}
  \end{minipage}
  \hfill
  \begin{minipage}[!t]{0.45\textwidth}
  \centering
  \caption{\textbf{Normalized Attention Maps} for parts which are localized (left) and non-localized (right).}
     \label{fig:attention-eval}
     \vspace{-2mm}
    \includegraphics[width=0.8\textwidth]{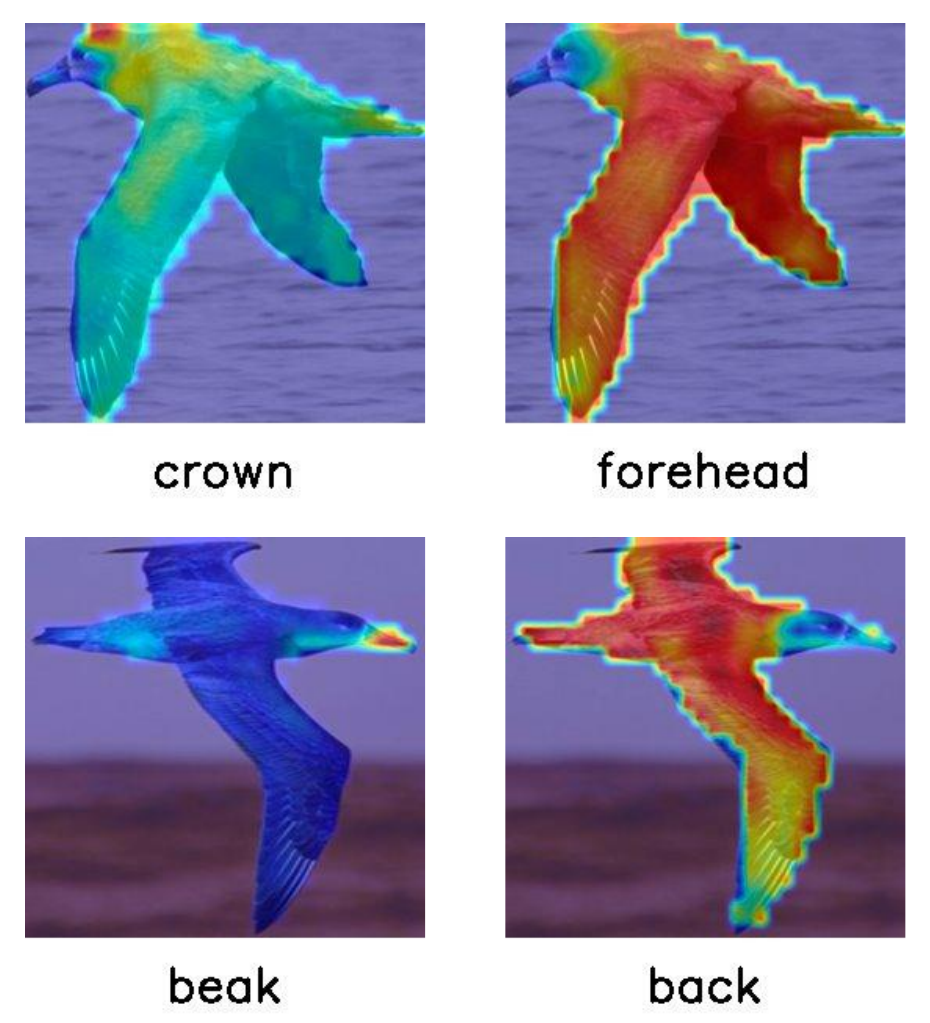}

  \end{minipage}
  \vspace{-5mm}
\end{figure*}
We use StableDiffusion (SD) 1.5 to compare the generation results to the Rich-Text~\cite{ge2023expressive}\footnote{https://github.com/SongweiGe/rich-text-to-image} baseline. In this case, we want to start the localization process at a later stage of denoising, as we aimed the part diffusion not to alter the base generation too much; hence, start part denoising from step $T_{th} = 24$. We do merging of base with initial diffusion process for $0.5$ number of steps (Eq.~\ref{eq:self_attn_inj})  and $\delta = 0.3$ for max part localization (in Eq.~\ref{eq:mask_thresh}) respectively. We use a PNDM scheduler with 41 steps and an 8.5 guidance scale, as done by default in Rich-Text~\cite{ge2023expressive} generation.

\vspace{1mm} \noindent \textbf{Part Generation.} We use the same scheduler and guidance scale for the part generation. We blend the base $x_t^{base}$ and part generations $x_t$ 
for 0.2 fraction of the time steps in the denoising process.

\vspace{1mm} \noindent \textbf{Baselines.} For the Rich-Text baseline, we use the attribute properties $\ba_i$ for the part token and add it to the base token if it is in the base prompt. In other cases, we add the Part information as the footnote of the object token in the base prompt. For the InstructPix2Pix~\cite{brooks2022instructpix2pix} baseline, we iteratively add the instructions for each part on the base generation through. For the StableDiffusion baseline, we add the PartComposer instructions as the object's description in the prompt. We use the same StableDiffusion model with the same seed and guidance scale across baselines. We run it on the same Nvidia A100 40GB to ensure the sanity of comparison across methods. PartCraft~\cite{ng2024partcraft}, a recent method, also aims to generate objects based on part composition. However, if the text-to-image model is PartCraft trained for CUB-200, it can only generate variations of CUB-200.
Further, it also requires annotations of the parts regarding key points, etc. Hence, it cannot generate parts based on text descriptions like our work. Hence, due to additional supervision and focus on particular datasets, it cannot be used as a general baseline for comparison with our work.

\section{Analysis of PartComposer}
In this section, we provide additional analysis regarding the design choices made in PartComposer. In particular, we ablate the components we have introduced in the PartDiffusion process of PartComposer (Sec.~\ref{subsec:part_loc}). We provide an analysis of the effect of using \textbf{a)} independent text embeddings, \textbf{b)} usage of dot product-based protocol in the part assignment, and \textbf{c)} the visualization of attention maps for the localized and unlocalized parts. We have used a subset of CUB-200 images to perform all evaluations, which is kept fixed across ablations. We tabulate the ablations in Table~\ref{tab:partcraft-ablation} (also including ablations from the main text Table~\ref{tab:partcraft-ablation-main}) and provide visual results in Fig.~\ref{fig:attention-eval}. We observe that all components introduced in PartComposer significantly contribute to the overall segmentation performance of the part localization module. Further, in Fig.~\ref{fig:attention-eval}, we find that normalizing parts that do not satisfy localization conditions (forehead and back) leads to high attention values in most regions. This demonstrates the effectiveness of the max-value-based selection (Eq.~\ref{eq:part_gen}) of parts proposed in PartComposer.
\begin{figure*}[!h]
    \centering
    \includegraphics[width=0.8\textwidth]{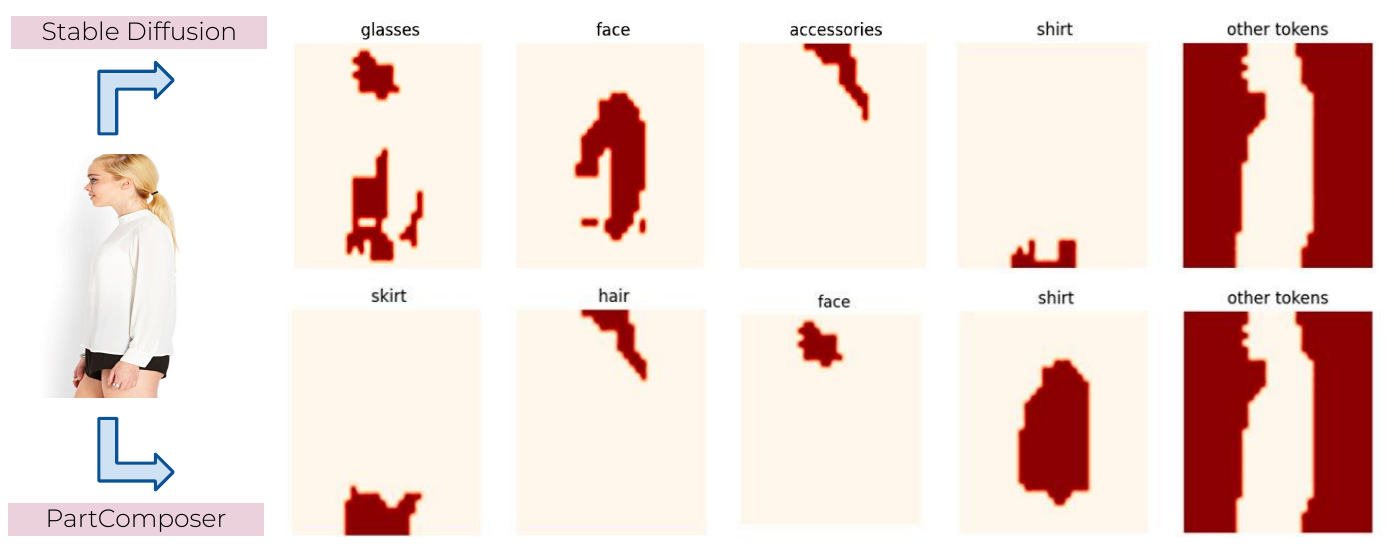}
    \caption{\textbf{Qualitative Comparison} of part masks (on DeepFashion) generated from the Stable Diffusion Baseline to PartComposer (Ours). The Stable Diffusion Baseline assigns arbitrary part masks to segments, whereas our part marks are consistent to part.}
    \label{fig:qual-masks-human}
\end{figure*}

\begin{figure}[!t]
    \centering
    \includegraphics[width=\columnwidth]{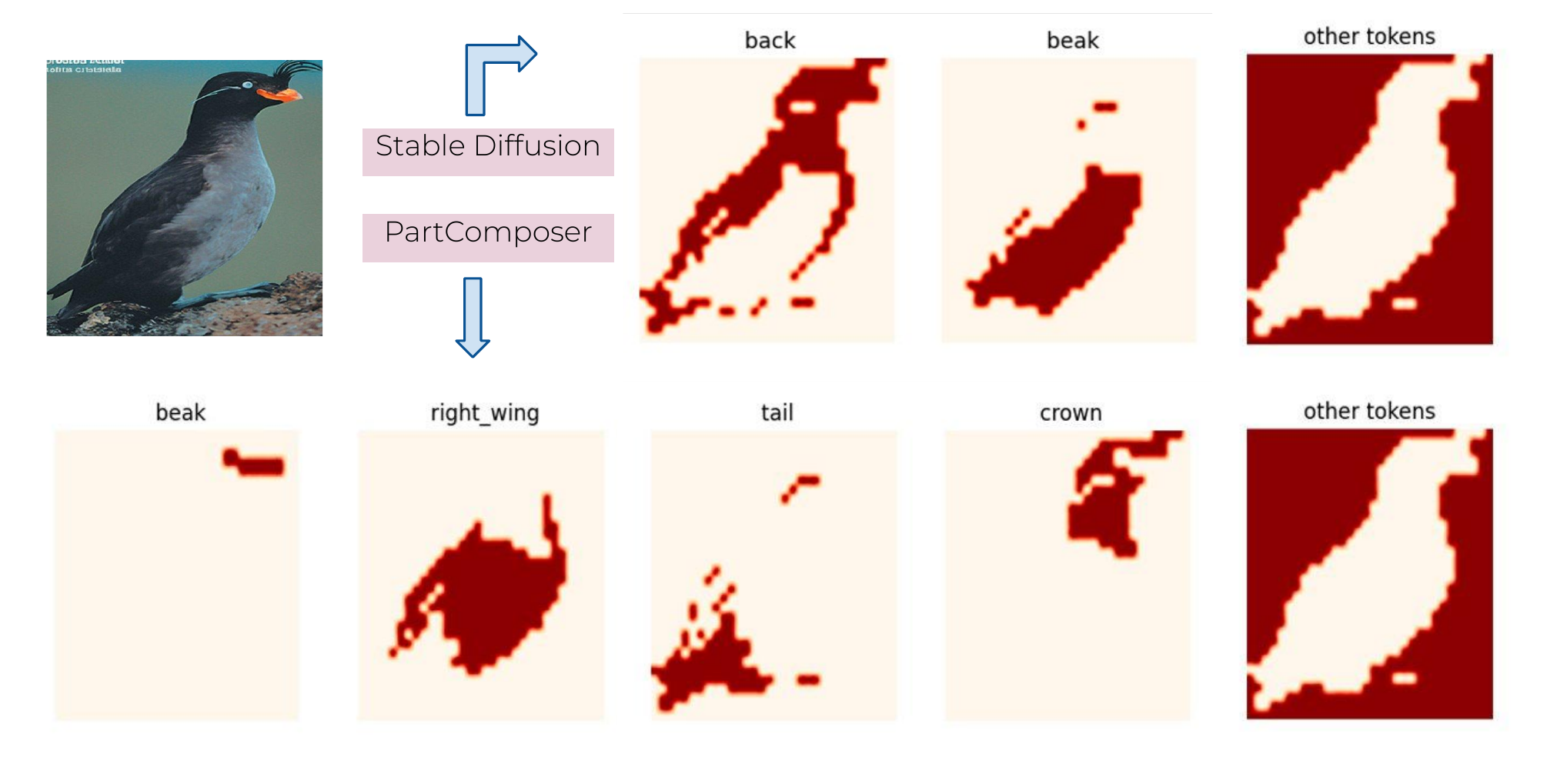}
    \caption{\textbf{Qualitative Comparison} of part masks (on CUB200) generated from the Stable Diffusion Baseline to PartComposer (Ours). The Stable Diffusion Baseline assigns arbitrary part masks to segments, whereas our part marks are consistent to part.}
    \label{fig:qual-masks-cub}
\end{figure}

For comparison to the Stable Diffusion baseline, in addition to the results provided in Table~\ref{tab:cub_deepfashion}, we provide a qualitative comparison in Fig.~\ref{fig:qual-masks-human} and~\ref{fig:qual-masks-cub}. The Stable Diffusion baseline assigns arbitrary masks to the wrong parts. On the contrary, if PartComposer localizes parts, they are often correctly associated with the right region of the object. This shows the advantage of using part diffusion rather than inducing additional tokens in the base prompt of StableDiffusion.

\begin{figure}[!h]
    \centering
    \includegraphics[width=\columnwidth]{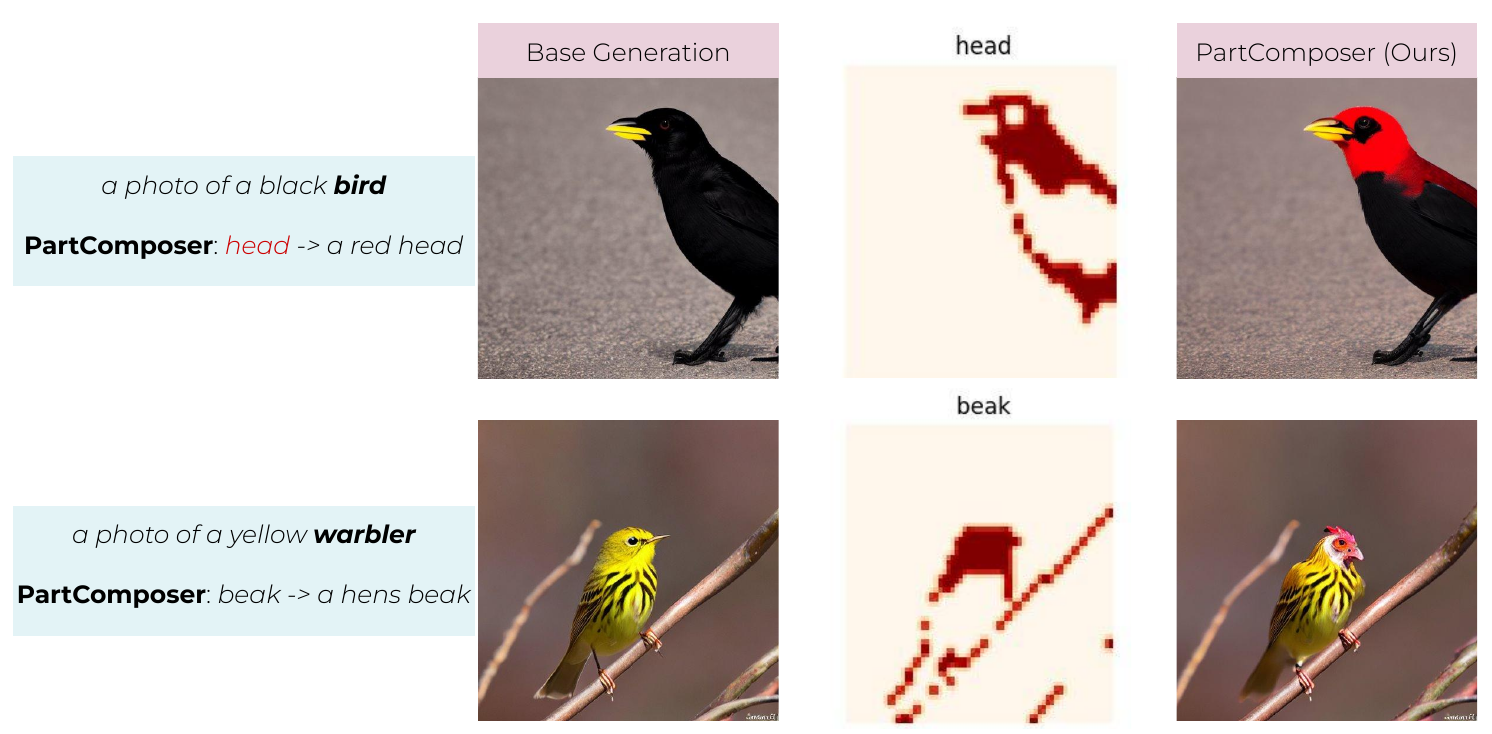}
    \caption{\textbf{PartComposer is robust when the masks don't exactly fit} the specified parts. This is because the text description for the masked region contains a description only of specified parts (\eg beak and head above). Hence, the other regions, despite being in the masked region, remain similar to the base generation. }
    \label{fig:part-craft-mask-robust}
\end{figure}

\vspace{1mm} \noindent \textbf{Robustnes of PartComposer w.r.t. Masks.} PartComposer uses masked diffusion process to generate and compose the object parts. Hence, we first see the effect of the mask region, where we observe (Fig.~\ref{fig:part-craft-mask-robust}) that in cases where the mask occupies more region than the desired part, the part diffusion process mostly modifies the requested part. This demonstrates that
PartComposer can still generate desired aesthetic outputs in case the masks
are not very accurate.

\null \newpage
\null \newpage

\section{Limitations of PartComposer Generations}

  \begin{figure}[!t]
        \centering
    \includegraphics[width=\columnwidth]{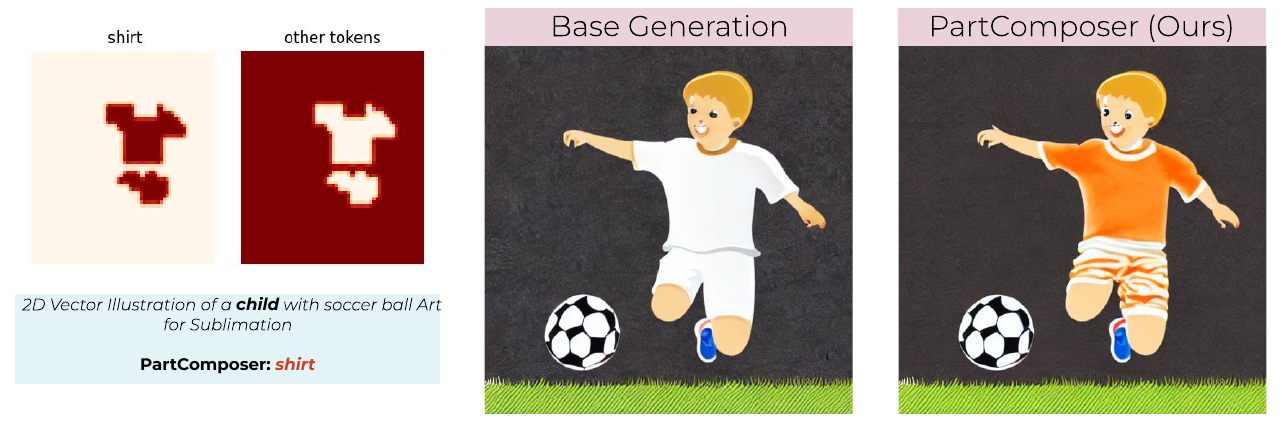}
    \caption{\textbf{ Limiting Case.} When only color is specified in PartComposer, and the masks cover extraneous regions. In such cases, when only color guidance is applied, it might leak to some other parts instead of the specified part (`shirt'). As color is added through gradient guidance in the entire masked region. }
    \label{fig:limiting-results}
\end{figure}

 \vspace{1mm} \noindent We find that the color guidance applies to the entire masked region. Hence, in some cases, the font color specification $\ba_{c}$, can alter other areas in minor ways besides the specified part. This is due to color gradient loss, also used in Rich-Text~\cite{ge2023expressive} to specify the specific color of the region. We highlight that in the example in Fig.~\ref{fig:limiting-results}, where some part of the orange lines appear in the shorts worn by the child as it was also the part of the mask. We find that mask localization is the bottleneck for performance. Hence, improving the part understanding and segmentation capability of the Text-to-Image models is an important direction to be pursued across future works, including explorations on new models like Flux and SD3.5.